\documentclass[11pt]{article}
\usepackage[a4paper, total={6in, 8in}]{geometry}

\usepackage[utf8]{inputenc} %
\usepackage[T1]{fontenc}    %
\usepackage{url}            %
\usepackage{booktabs}       %
\usepackage{amsfonts}       %
\usepackage{nicefrac}       %
\usepackage{microtype}      %
\usepackage{natbib}

\usepackage{csquotes} %quotes
\usepackage{bm} % bold symbols
\usepackage{amsmath, amssymb, amsfonts}
\usepackage{dsfont}
\usepackage{stmaryrd}
\newcommand{\bx}{\mathbf{x}}
\newcommand{\by}{\mathbf{y}}
\newcommand{\bth}{\bm{\theta}}
\newcommand{\bu}{\mathbf{u}}
\newcommand{\bz}{\mathbf{z}}
\newcommand{\bs}{\mathbf{s}}

\renewcommand{\mid}{\,|\,}

\usepackage{amsthm}
\theoremstyle{definition}
\newtheorem{remark}{Remark}
\newtheorem{definition}{Definition}

\usepackage[acronym,nowarn,section,nonumberlist]{glossaries}
\glsdisablehyper{}
\newacronym{sbi}{SBI}{simulation-based inference}
\newacronym{lfi}{LFI}{likelihood-free inference}
\newacronym{bsl}{BSL}{Bayesian synthetic likelihood}
\newacronym{snre}{SNRE}{sequential neural ratio estimation}
\newacronym{snpe}{SNPE}{sequential neural posterior estimation}
\newacronym{snle}{SNLE}{sequential neural likelihood estimation}
\newacronym{nre}{NRE}{neural ratio estimation}
\newacronym{npe}{NPE}{neural posterior estimation}
\newacronym{nle}{NLE}{neural likelihood estimation}
\newacronym{abc}{ABC}{approximate Bayesian computation}
\newacronym{mlp}{MLP}{multi-layer perceptron}
\newacronym{sam}{SAM}{sharpness-aware minimisation}
\newacronym{elbo}{ELBO}{evidence lower-bound}
\newacronym{tg}{TG}{toy Gaussian}
\newacronym{tgss}{TG-SS}{toy Gaussian with summary statistics}
\newacronym{slcp}{SLCP}{simple likelihood complex posterior}
\newacronym{sv}{SV}{stochastic volatility}
\newacronym{svss}{SV-SS}{stochastic volatility with summary statistics}
\newacronym{ood}{OOD}{out-of-distribution}
\newacronym{gbi}{GBI}{generalised Bayesian inference}
\newacronym{dgp}{DGP}{data-generating process}

\usepackage{subcaption}
\usepackage{float}

\usepackage[usenames,dvipsnames]{xcolor}
\definecolor{shadecolor}{gray}{0.9}

\usepackage{graphicx}
\usepackage[labelfont=bf]{caption}
\usepackage{subcaption}

\usepackage{booktabs,array}

\usepackage[algoruled]{algorithm2e}
\usepackage{listings}
\usepackage{fancyvrb}
\fvset{fontsize=\small}

\usepackage[
    pdftitle={Investigating the Impact of Model Misspecification in Simulation-based Inference},    % title
    pdfauthor={Patrick Cannon, Daniel Ward, Sebastian M Schmon},     % author
    colorlinks,
    linktoc=all]
{hyperref}
\hypersetup{allcolors=MidnightBlue}

\usepackage{todonotes}
\usepackage{wrapfig}
\usepackage{tcolorbox}

\usepackage[affil-it]{authblk}
\title{Investigating the Impact of Model Misspecification in\\ Neural Simulation-based Inference}
\author[1]{Patrick Cannon\footnote{
\href{mailto:patrickcannon@improbable.io}{patrickcannon@improbable.io} 
}}
\author[2]{Daniel Ward}
\author[1,3]{Sebastian M. Schmon}
\affil[1]{Improbable, UK}
\affil[2]{School of Mathematics, Bristol University, UK}
\affil[3]{Department of Mathematical Sciences, Durham University, UK}
\date{}                     %
\begin{document}

\maketitle

\begin{abstract}
Aided by advances in neural density estimation, considerable progress has been made in recent years towards a suite of \gls{sbi} methods capable of performing flexible, black-box, approximate Bayesian inference for stochastic simulation models.
While it has been demonstrated that neural \gls{sbi} methods can provide accurate posterior approximations, the simulation studies establishing these results have considered only \emph{well-specified} problems -- that is, where the model and the data generating process coincide exactly. However, the behaviour of such algorithms in the case of model \emph{misspecification} has received little attention.
In this work, we provide the first comprehensive study of the behaviour of neural \gls{sbi} algorithms in the presence of various forms of model misspecification. 
We find that misspecification can have a profoundly deleterious effect on performance.
Some mitigation strategies are explored, but no approach tested prevents failure in all cases. 
We conclude that new approaches are required to address model misspecification if neural \gls{sbi} algorithms are to be relied upon to derive accurate scientific conclusions. 
\end{abstract}

\section{Introduction}
\glsreset{sbi}
Across the sciences, systems of interest are routinely modelled using stochastic simulations. They offer domain experts the capacity to describe intuitively their understanding of the mechanistic behaviour under study, free from the constraints of analytic tractability.
Though simulators can offer incredible flexibility and emergent complexity to modellers, by the same token they are rarely amenable to traditional statistical inference techniques. By definition, they are easily run forward, mapping input parameters $\bth$ to samples $\bx$, implicitly defining a likelihood $p(\bx \mid \bth)$, but this function is generally intractable. 
Analytic Bayesian inference of the parameters is therefore ruled out since, after specifying a prior distribution $\pi(\bth)$, inference centres on the intractable posterior distribution $\pi(\bth \mid \bx) \propto p(\bx \mid \bth) \pi(\bth)$. 

Substantial effort has gone toward developing so-called \gls{sbi} methods, which use model samples to approximate the posterior distribution in settings where the likelihood function is intractable \citep{cranmer2020frontier}. Perhaps the most established of these is \gls{abc}, a conceptually simple though powerful Monte Carlo technique; \cite{beaumont2019review} provides a review. 
More recently, a branch of research has emerged taking advantage of neural density estimation methods, such as normalising flows \citep{papamakarios2021normalizing}, to approximate either the likelihood or posterior distribution directly \cite[see e.g.][]{lueckmann2021benchmarking}. Such neural \gls{sbi} techniques benefit from the flexibility, efficiency and amortisation made possible by recent developments in probabilistic generative modelling and deep learning as a whole, and are now widely used, for example in epidemiology \citep[][]{ARNST2022108805}, astrophysics \citep[][]{green2020}, and economics \citep{dyer2022black}.

Tempering these benefits, however, is a crucial shortcoming: the behaviour of neural \gls{sbi} methods under misspecification is poorly understood. If a discrepancy exists between the simulator and the real data-generating process, then the performance of the algorithms will depend on their ability to generalise to out-of-distribution data. It is well known that many generative models in machine learning, such as variational autoencoders and normalising flows, generalise poorly to out-of-distribution data \citep{nalisnick2018do}, which suggests some cause for concern in downstream applications like \gls{sbi}.
A figurative alarm has already been raised in \cite{hermans2021averting}, which demonstrates imperfect calibration of approximate posteriors found using virtually all popular neural \gls{sbi} techniques, fundamentally calling into question their reliability even in well-specified settings. This paper corroborates and extends this line of inquiry to misspecified models, which we argue is by some margin the most common setting in which these techniques are used.

The following contributions are made:
\begin{enumerate}
    \item We carry out the first benchmark of neural \gls{sbi} methods under misspecification, extending the work of \cite{lueckmann2021benchmarking} and \cite{hermans2021averting}.
    \item We show that the na\"ive use of the most common classes of density estimators is currently untenable in situations where any appreciable level of model error is present.
    \item We investigate mitigation strategies to improve the robustness of neural \gls{sbi} methods to misspecification. We demonstrate the utility of ensemble posteriors, as well as training procedures like \gls{sam}, but find that neither is a panacea.
\end{enumerate}

Greater understanding of \gls{sbi} methods is, we believe, urgent. While computational modelling has for many decades influenced public policy, the role of epidemiological models in the global response to the COVID-19 pandemic \citep[e.g.][]{ferguson2020impact, spooner2021dynamic} has put into sharp relief the necessity of defensible scientific reasoning. It is in exactly such situations, where the system under study is extremely complex and the decisions to be made high-stakes, that we are likely to find in tension the need to use intractable models but to understand them well. In other words, the highest standards of accuracy are required in settings where it is extremely difficult to achieve them. We hope that practitioners and researchers of \gls{sbi} will embrace this responsibility and work to resolve the shortcomings presented here, else the methodologies can never be used where they are needed.

\section{Problem Statement}
\subsection{Model Misspecification}\label{section:model_misspecification}

A computer simulation with parameters $\bth \in \Theta \subset \mathbb{R}^d$ defines implicitly a parametric family of distributions
\begin{equation}
    \mathcal{P} \triangleq \{ p(\cdot \mid \bth); \bth \in \Theta \}.
\end{equation}
Running the simulator conditional on a parameter $\bth \in \Theta$ is equivalent probabilistically to sampling an output $\bx$ from the likelihood $p(\bx \mid \bth)$.
 In a typical scientific modelling setting, the researcher has access to one or more independent observations $\by$, drawn from an unknown data-generating process $p^*$, as well as the simulator family $\mathcal{P}$ which it is assumed offers a reasonable explanation for the data.
In the best case scenario, $p^* \in \mathcal{P}$, so that for some $\bth^*$, $p^* = p(\cdot \mid \bth^*)$ and the model family $\mathcal{P}$ is said to be \emph{well-specified} with respect to $\by$. If on the other hand $p^* \notin \mathcal{P}$, then $\mathcal{P}$ (or any model in it) is said to be \emph{misspecified} with respect to $\by$. In either scenario, provided $\by$ lies in the support
\begin{equation}
    \mathcal{X} \triangleq \big\{ \bx: \textstyle \int p(\bx \mid \bth)\pi(\bth) \, \mathrm{d}\bth > 0 \big\}
\end{equation}
of the model evidence, the posterior distribution $p(\bth \mid \by)$ is defined. Evaluating this posterior distribution is the foundation of Bayesian statistical inference in the sciences. Unfortunately, models defined by computer simulations in general do not enjoy an analytically tractable likelihood function, and therefore the corresponding posterior must be found through alternative means. 
An emerging literature has proposed a variety of approaches to achieve this by estimating one of: \emph{1)} the posterior \citep[e.g.][]{tavare97, beaumont2009adaptive, papamakarios2016fast,greenberg2019automatic}, \emph{2)} the likelihood \citep[e.g.][]{diggle1984monte, wood2010statistical, grazzini2017bayesian, papamakarios2019sequential}, or \emph{3)} the likelihood-to-evidence ratio \citep[e.g.][]{thomas2022likelihood, durkan2020contrastive, hermans2020likelihood, dyer2022amortised}.

Many of the more recent developments make use of neural networks as flexible function approximators. Throughout this work we will concentrate on these approaches, in particular \gls{npe}, \gls{nle} and \gls{nre}, which we term collectively \emph{neural} \gls{sbi}. Whatever the neural \gls{sbi} method employed, the outcome of training is a function $q(\bth \mid \bx):\Theta \times \mathcal{X} \rightarrow \mathbb{R}^+$ approximating (perhaps up to a constant) the model posterior $p(\bth \mid \bx)$. These methods have been shown to recover the Bayesian posterior distribution $p(\bth \mid \bx)$ with reasonable accuracy when $\bx$ is sampled from the model \citep{lueckmann2021benchmarking}.
In this work, we investigate the accuracy of those methods when this assumption is not fulfilled. 
\begin{figure}
  \centering
  \includegraphics[width=.99\linewidth]{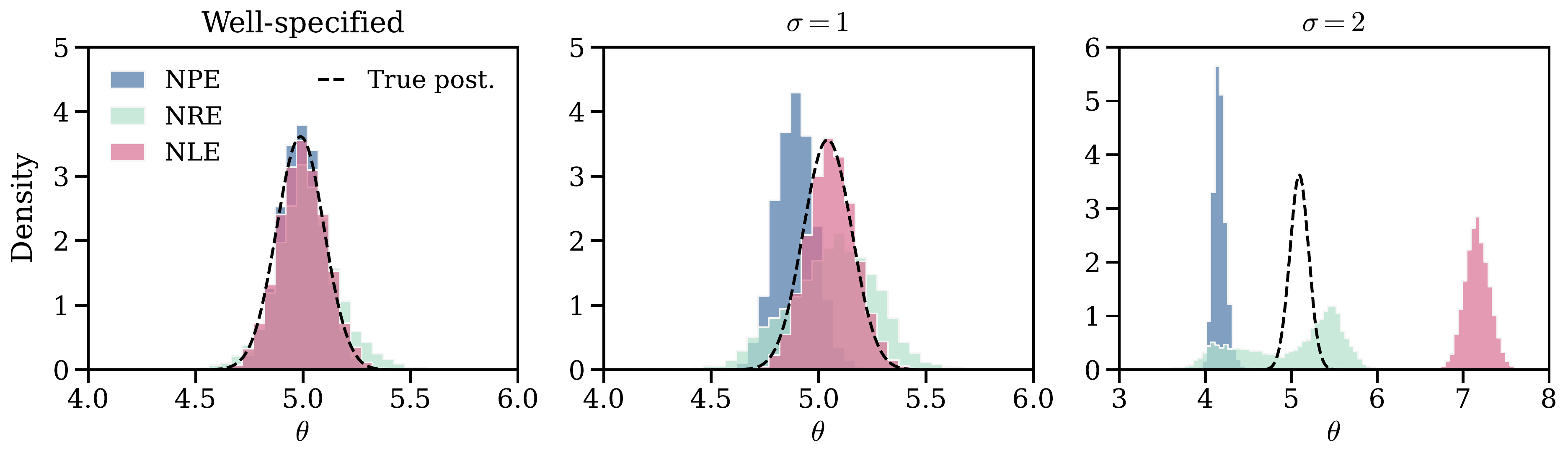}  
\caption{Approximate posterior distributions derived using three \gls{sbi} algorithms, in the well-specified regime and under two levels of data misspecification. The dashed line shows the ground-truth model posterior $p(\bth \mid \by_\sigma)$. Taking $\sigma=0$ gives $\by_0=\bx$, a sample from the model itself.}
\label{fig:toygauss_introduction}
\end{figure}

The following toy problem \citep[taken from][]{frazier2017model} demonstrates clearly the egregious impact of misspecification on neural \gls{sbi} techniques. Using a one-dimensional parameter $\bth$, with prior $\bth \sim \mathcal{N}(0,5^2)$, a simulator is defined by taking the mean and the standard deviation of 100 independent samples from a Gaussian distribution, i.e. $\bx = s(\bu)$ where $s(\bu) = (\texttt{mean}(\bu), \texttt{sd}(\bu))$ and $\bu \sim \mathcal{N}(\bth; I_{100})$. 
The observation, $\by_\sigma$, will come from a similar model, but where we are able to vary the standard deviation of the \emph{iid} normal random variables. 
In practice, we add zero-mean Gaussian noise to the model output to achieve this: $\by_\sigma = s(\bu + \sigma\bz)$ with $\bz \sim \mathcal{N}(0,I_{100})$.
For $\sigma = 0$, the model is well-specified. For any value $\sigma \neq 0$, the model will be misspecified with respect to $\by_\sigma$.
The results can be seen in Figure \ref{fig:toygauss_introduction}. All methods closely approximate the true posterior when no misspecification is present, but performance deteriorates drastically with increasing levels of misspecifcation. For $\sigma = 2$ the approximate posteriors become unusable; all are highly inaccurate in different ways, with \gls{nle} and \gls{npe} covering entirely disjoint areas of the parameter space. This behaviour is consistent, in the sense that over different training seeds a similar decline in accuracy is always observed, but inconsistent in the sense that the decline in accuracy of any particular instantiation of each algorithm is unpredictable and so could not be systematically accounted for by explicitly modelling the bias.

In summary, the neural density estimators shown provide inaccurate approximations of the model posterior $p(\bth \mid \by)$ because they fail to generalise well to points $\by$ in the tails of their training data, that is, points in the tails of the model marginal likelihood $p(\bx)$. This inaccuracy is rarely evident when evaluating approximations $q(\bth \mid \by)$ for observations $\by$ drawn from the model because by definition such points rarely fall in the tails of the distribution they come from. In contrast, when considering the approximation $q(\bth \mid \by)$ when $\by \sim p^*$, it is entirely plausible for $\by$ to lie in the tails of $p(\bx)$ and therefore to be highly anomalous with respect to the training data.

\subsection{The Aim of this Study}

Before discussing potential strategies for mitigating the impact of model misspecification, we define clearly the problem we seek to address.
The discrepancy between the induced distribution of the simulator and that of the observed data, $\by$, has found increasing attention in the study of classical approaches to \gls{sbi} such as \gls{abc} \citep[e.g.][]{frazierrobust, frazier2017model}, \gls{bsl} \citep[e.g.][]{frazier2021synthetic, frazier2021robust} and \gls{gbi} \citep[e.g.][]{schmon2020generalized} as well as related approaches \citep{dellaporta22a}. However, so far unexplored is how \emph{neural} \gls{sbi} techniques perform when the observation data, $\by$, is not compatible with the induced simulator distribution $p(\bx \mid \bth)$ and whether such techniques retain their advantages highlighted in previous benchmarks for well-specified models \citep{lueckmann2021benchmarking}.
We therefore aim to answer the following question:
\begin{tcolorbox}[colback=YellowGreen!40]
Do neural SBI methods offer a good approximation to the model posterior even when the data-generating process does not coincide with the model?
\end{tcolorbox}

In the remainder of this paper, we will investigate this behaviour across a variety of tasks with artificial misspecification of different types. Before we describe our experiments, we first discuss several potential strategies to reduce the likelihood of failure under misspecification.

\begin{remark}[Open problems]
The application of \gls{sbi} in a misspecified setting gives rise to many potential lines of scientific enquiry.
For example: if model misspefication is suspected, is it even desirable to target the model posterior? Is the use of an alternative, `robust', posterior, as is done in \gls{gbi}, to be preferred? Under what conditions does a posterior distribution for the true data-generating process exist? These questions are currently without definitive answers and addressing them is sure to be at the core of forthcoming research.
However, we stress that they are not the focus of this work.
\end{remark}

\begin{remark}[Sequential methods]
The methods we discuss here are often used in a sequential fashion, taking advantage of active learning principles to increase efficiency \citep[see e.g.][]{hermans2020likelihood, papamakarios2019sequential}. In particular, over a number of `rounds' of training, intermediate posteriors are constructed which then serve as the proposal distribution in a subsequent round. This approach can easily exacerbate the degeneracies observed in Figure \ref{fig:toygauss_introduction}; if an intermediate posterior approximation is poor, subsequent rounds are highly unlikely to offer a substantial improvement. For this reason we do not consider them in this work.
\end{remark}

\subsection{Robustifying Neural Network Approximations}\label{sec:mitigation-strategies}

We now describe two mitigation strategies that will be systematically tested across a range of tasks in Section \ref{sec:experiments}. Both strategies attempt to address what we see as the core issue affecting the performance of neural \gls{sbi} algorithms in misspecified settings: \gls{ood} performance of neural networks. 

Despite their state-of-the-art predictive performance across a wide range of tasks, neural networks are known to produce overconfident predictions and offer little in the way of predictive uncertainty quantification \citep[][]{lakshminarayanan2017simple}. This problem is particularly acute in the case of \gls{ood} data, that is, data drawn from a distribution that is distinct from that of the training data. This is exactly the setting of the current work, where the training data for \gls{sbi} algorithms is drawn from the model, $(\bth, \bx) \sim \pi(\bth)p(\bx \mid \bth)$, but the real data comes from an unknown data-generating process $\by \sim p^*$ with potentially starkly different geometry to the model.

\paragraph{Ensemble Posteriors.} 
An ensemble Bayesian posterior is simply a mixture of independent posterior distributions. In the current setting, the ensemble of interest is 
\begin{equation}
    \bar{q}(\bth \mid \bx) \triangleq \frac{1}{n} \sum_{i=1}^n q_{\phi_i}(\bth \mid \bx)
\end{equation}
where each of the $q_{\phi_i}(\bth \mid \bx)$ is the result of the standard training procedure for the respective algorithm, trained using a distinct random seed, with $\phi_i$ denoting the particular local minima in parameter space. The idea is intuitive, appealing to a ``wisdom of the crowd'' logic, and has been used to improve neural network generalisability to unseen data \citep[see e.g.][]{wilson2020bayesian}. It is a particularly appealing strategy for neural \gls{sbi} posterior approximations since, as demonstrated in Figure \ref{fig:toygauss_introduction}, they can exhibit individually catastrophic inaccuracy. \cite{hermans2021averting} considered ensembles in the well-specified case, finding that they offer greater reliability, though may still be over-confident relative to the true posterior distribution. Ensembles feature prominently in deep learning more widely, where they have been used to improve robustness and generalisability \citep[][]{lakshminarayanan2017simple}. In Bayesian deep learning, ensembles are often invoked implicitly through use of posterior predictive samples from the neural network.

Ensembles are related to \emph{bagging}, an approach which uses an ensemble of posteriors conditioned on bootstrapped subsets of the data, which was also explored in \cite{hermans2021averting}. We note that bagging is a promising approach to tackling model misspecification \citep{huggins2019robust}, and indeed is efficiently achieved for single-round neural \gls{sbi} posteriors via amortisation, but do not pursue this approach here as a naive use of bagging is only sensible for \emph{iid} data. 

\paragraph{Robust Optimisation.} 
\glsreset{sam}
Deep learning algorithms make substantial use of overparameterisation. As a consequence, when trained using gradient descent on a simple training-set loss they can readily overfit the training data. This manifests as poor generalisation to unseen data. Their \emph{loss landscape}, a term to describe the surface of the loss function on which they are trained, can exhibit highly unusual geometry, in particular many sharp minima. There is a known connection, however, between the flatness of (the neighbourhood around) the minima, and the generalisation ability of the resulting model \citep[e.g.][]{keskar2016large}. Several techniques have been proposed to find improved areas of the parameter space, for example stochastic weight-averaging \citep{izmailov2018averaging} and a parametric Gaussian variation \citep{maddox2019simple}. Both techniques offer a way of considering iterates of stochastic gradient descent as samples from the posterior distribution of a Bayesian neural network, providing some uncertainty quantification without requiring substantial additional code for existing models. In this paper we explore the application of \gls{sam} \citep{foret2021sharpnessaware} -- a training algorithm that allows for efficient discovery of these `flat' minima, and is capable of improving generalisation in several classical deep learning tasks. We test this approach in Section \ref{sec:experiments} to determine whether finding flatter regions of the loss landscape improves the chances of the neural \gls{sbi} density estimators generalising to out-of-distribution data.

\section{Experiments}\label{sec:experiments}

\subsection{Setup}
In this section we consider three example models, defined by a parameter prior distribution $\pi(\bth)$ and a simulator $p(\bx \mid \bth)$. To facilitate performance benchmarking, each model was chosen because there exists an exact method for sampling from the posterior distribution $p(\bth \mid \bx)$.
For each task considered, we generate a set of observations and transform them using five degrees of increasing misspecification from the model. 
To achieve this we first take a sample $(\bth, \bx) \sim \pi(\bth)p(\bx \mid \bth)$ from the joint distribution of the model. Next, to emulate model misspecification, we perturb the observation $\bx$ using a (potentially stochastic) \emph{misspecification transform} $T_{\sigma}$. We define it in such a way that $T_0$ is the identity map, and so that the degree of misspecification of the data-generating process from the model increases in $\sigma$.
In particular, we employ a series of five misspecification transforms $T_{\sigma}, \sigma \in \llbracket 0,4 \rrbracket$. This staged sequence of misspecification allows us to perform a more nuanced analysis, observing the effect of both small and large deviations from the model assumptions. In some cases, $T_\sigma$ depends on an auxiliary random variable $\bz$, in which case the transformation of the observation $\bx$ is written $T_\sigma(\bx ; \bz)$.

To ensure that our results are not due to an unfortunate choice of $\bx$, the misspecification transform is applied to 50 \emph{iid} samples $(\bth_i, \bx_i) \sim \pi(\bth)p(\bx \mid \bth)$.
The result is a $50 \times 5$ collection of observations $(T_{\sigma}(\bx_i; \bz_i))_{i,\sigma \in \llbracket 1,50 \rrbracket \times  \llbracket 0, 4 \rrbracket}$ for each model. Our experiments probe the accuracy of the approximation
\begin{equation}
\label{eq:posterior_approximation}
    q(\bth \mid T_{\sigma}(\bx_i; \bz_i)) \approx p(\bth \mid T_{\sigma}(\bx_i; \bz_i))
\end{equation}
across $\sigma$, for each model and each neural \gls{sbi} approach using metrics we describe in the sequel.

As in Section \ref{section:model_misspecification}, we investigate the performance of three broad classes of neural \gls{sbi} methods (\gls{nre}, \gls{nle} and \gls{npe}) as implemented in the \texttt{sbi} package \citep{tejero-cantero2020sbi} to find the posterior approximation \eqref{eq:posterior_approximation}. As a baseline, each approach is compared to a simple rejection \gls{abc} algorithm, which always uses $10^5$ model samples, retaining the best 1\%.

For the flow-based methods, \gls{npe} and \gls{nle}, we use neural spline flows \citep[][]{durkan2019neural}, and for the classifier-based \gls{nre} we use a simple fully connected neural network. See the supplementary material for further implementation details.

\begin{wrapfigure}{r}{0.32\textwidth}
\vspace{-6mm}
  \begin{center}
    \includegraphics[width=\linewidth]{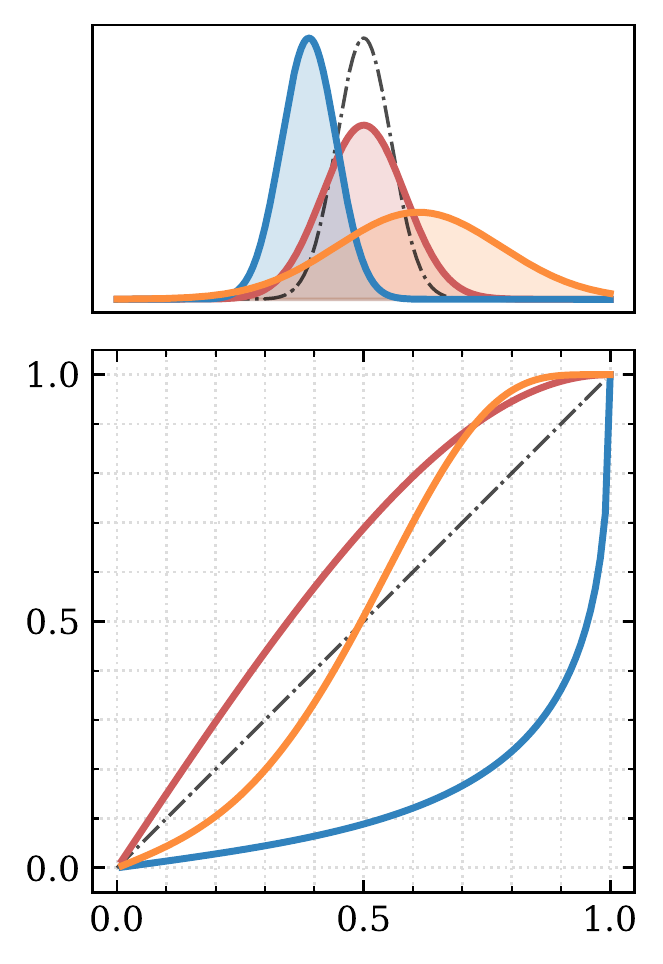}
  \end{center}
  \vspace{-5mm}
  \caption{Visualising coverage: three exemplar density functions (coloured) and their coverage of the base distribution (dash-dotted).\vspace{-10mm}}
  \label{fig:coverage_example}
\end{wrapfigure}

While the following experiments are presented for a particular choice of flows and neural networks, we found in experiments that neither architectural details of the neural networks (e.g. number of hidden units, layers) nor the flows (e.g. choice of transform) changed the qualitative outcome of the analysis.

Finally, we note that the misspecification transforms have been intentionally chosen to be simple. For example, the noise is never parameter dependent, and in two examples the misspecification is simply additive Gaussian noise. Despite this, we shall see that the methods tested still struggle greatly in this setting.

\subsection{Metrics}
To assess accuracy in the experiments, we consider the \emph{coverage} of the posterior approximations. As discussed in \cite{hermans2021averting}, coverage usefully describes the kind of performance that is important for defensible scientific inference, in the sense that it not only assesses the accuracy of the posterior approximation but can inform the researcher at a glance if the approximation is conservative or overconfident.

\begin{definition}[Expected Coverage]
Denote by $\Theta_{p(\bth \mid \bx)}(1-\alpha) \subset \Theta$ the $100(1-\alpha)\%$ highest posterior density region of $\Theta$ with respect to $p(\bth \mid \bx)$. That is, 
\begin{equation}
    \int_{\Theta_{p(\bth \mid \bx)}(1-\alpha)} p(\bth \mid \bx) \, \mathrm{d}\bth = 1-\alpha \quad \mathrm{and} \quad p(\bth \mid \bx) > p(\bth' \mid \bx) 
\end{equation}
for any $\bth \in \Theta_{p(\bth \mid \bx)}(1-\alpha)$, $\bth' \in \Theta \setminus \Theta_{p(\bth \mid \bx)}(1-\alpha)$ \citep[e.g.][]{mukhopadhyay2000probability}. 
Using the notation $h_{\alpha, p}(\bth, \bx) = \mathds{1}\{ \bth \in \Theta_{p(\bth \mid \bx)}(1-\alpha)\}$, the \emph{expected coverage} is defined to be $\mathbb{E}_{p(\bth, \bx)} [ h_{\alpha, p}(\bth, \bx) ]$ which is easily shown to be equal to $1-\alpha$. 
\end{definition}

In \cite{hermans2021averting}, the expectation $\mathbb{E}_{p(\bth, \bx)} [ h_{\alpha, p}(\bth, \bx) ]$ is approximated for a particular posterior estimator $q(\bth \mid \bx)$ using the Monte Carlo average 
\begin{equation}\label{eq:expected_coverage}
    \frac{1}{n} \sum_{i=1}^n h_{\alpha, q}(\bth_i, \bx_i); \quad \quad (\bth_i, \bx_i) \overset{iid}{\sim} \pi(\bth)p(\bx \mid \bth).
\end{equation}
For a well-calibrated estimator $q$, Equation \eqref{eq:expected_coverage} evaluates to $1-\alpha$ at any level $\alpha$. For this reason the performance of the estimator can be assessed by plotting the nominal and actual coverage against one another for a range of values $\alpha \in [0,1]$, with the line of equality ($45^\circ$ line) %
being optimal. An example is shown in Figure \ref{fig:coverage_example}.
Deviations above the straight line (actual coverage exceeding nominal coverage) indicate that the approximation $q$ is conservative at that level, with deviations below the line indicating over-confidence. 
To assess performance in the misspecified setting, this approach must be adjusted. To see this, notice that the true data-generating process could take the form 
\begin{equation}\label{misspecified_dgp}
    \by \sim \tilde{p}(\by) \triangleq \int \pi(\bth) \tilde{p}(\by \mid \bth) \, \mathrm{d}\bth \triangleq \int \pi(\bth) \int p(\bx \mid \bth) g(\by \mid \bx) \, \mathrm{d}\bx \, \mathrm{d}\bth
\end{equation}
where $g(\by \mid \bx)$ is some observation noise model. Analogously to the approach taken above, we might consider approximating 
\begin{equation}\label{eq:wrong_coverage}
    \mathbb{E}_{\tilde{p}(\bth, \by)} \left[ \mathds{1}\{ \bth \in \Theta_{q(\bth \mid \by)}(1-\alpha)\} \right]
\end{equation}
using Monte Carlo techniques as in \eqref{eq:expected_coverage}. Notice however that $\tilde{p}(\bth, \by) = \tilde{p}(\by) \tilde{p}(\bth \mid \by)$ and that therefore Equation \eqref{eq:wrong_coverage} features the inner expectation $\mathbb{E}_{\tilde{p}(\bth \mid \by)} \left[ \mathds{1}\{ \bth \in \Theta_{q(\bth \mid \by)}(1-\alpha)\} \right]$ which does not in general evaluate to $1-\alpha$ for a perfect approximation $q=p$ as it did in the well-specified case. Fortunately, an alternative approach is possible. Writing the expected coverage as the expected Bayesian credibility
\begin{equation}
    \mathbb{E}_{p(\bth, \by)} \left[ \mathds{1}\{ \bth \in \Theta_{p(\bth \mid \by)}(1-\alpha)\} \right] = \mathbb{E}_{p(\by)} \mathbb{E}_{p(\bth \mid \by)} \left[ \mathds{1}\{ \bth \in \Theta_{p(\bth \mid \by)}(1-\alpha)\} \right] = 1-\alpha
\end{equation}
reveals that, since it is the inner expectation that evaluates to $1-\alpha$, the outer expectation could be taken over any $\by$, in particular according to Equation \eqref{misspecified_dgp}.

\subsection{Task: Toy Gaussian Model}
\glsunset{tg}
\glsunset{tgss}
This is a simple Gaussian model, devised by \cite{frazier2017model}. Here, the model is given by $\bx=(\bx_1,\ldots, \bx_{100})$ with $\bx_i \sim^{iid} \mathcal{N}(\bth, 1)$ and parameter prior distribution $\bth \sim \mathcal{N}(0, 5^2)$. The data-generating process is taken to be $\by = (\by_1,\ldots, \by_{100})$ with $\by_i \sim^{iid} \mathcal{N}(\bth, r^2)$, with the same prior. It follows that for $r \neq 1$ the model is misspecified. 
For the misspecification transforms, we generate for each observation Gaussian noise variable $\bz$ and use $T_{\sigma}(\bx; \bz) = \bx + \sigma \bz$ where $\bz \sim \mathcal{N}(\bm{0}, I_{100})$
for $\sigma \in \{0,\ldots,4\}$. The level $\sigma=0$ represents no misspecification. We consider this model both with and without summary statistics which, when used, take the form $\bs(\bx)=(s_1(\bx), s_2(\bx))$ with $s_1(\cdot)$ and $s_2(\cdot)$ the sample mean and standard deviation respectively. Summary statistics, if present, are applied after the transform $T_{\sigma}$ is applied. For brevity we will refer to the standard version of the model as \gls{tg} and the version with summary statistics applied as \gls{tgss}. Reference posterior samples are easily obtained; the prior distribution and likelihood are conjugate so the posterior distribution is available in closed form.

\begin{figure}[!htb]
  \centering
  \includegraphics[width=\linewidth]{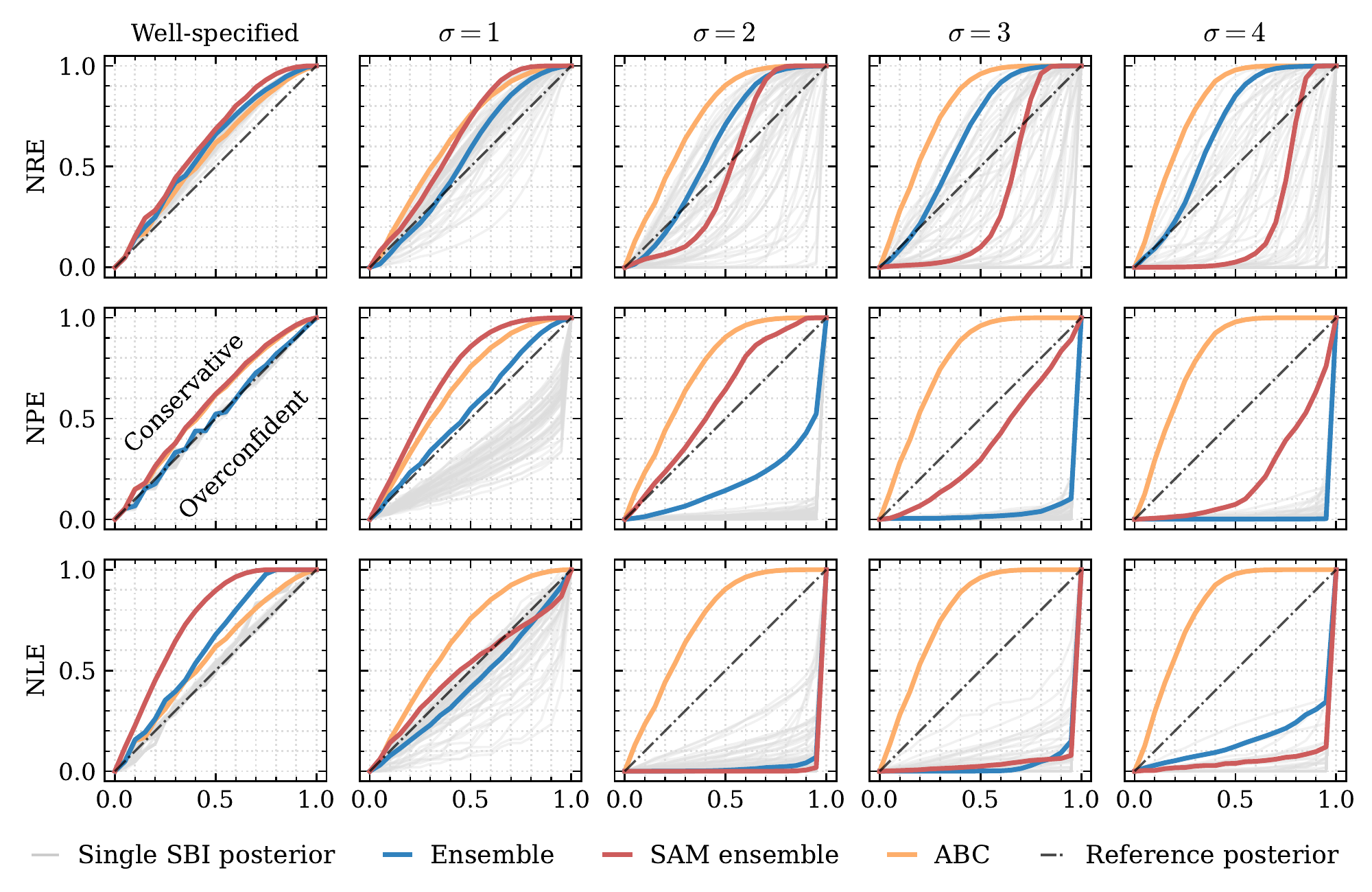}  
\caption{Coverage results for all algorithms on \gls{tgss} task with $10^5$ training samples.}
\label{fig:tg-condensed}
\end{figure}

\paragraph{Results.} Figure \ref{fig:tg-condensed} shows the coverage results for $10^5$ training samples using summary statistics. For this, and all other tasks, full results are found in the supplement.
 We observe that as the level of misspecification increases, the posterior approximations in general become less conservative and the variance in coverage over distinct posterior approximations increases. In other words, the more misspecified the model is for the data, the more unpredictable the behaviour of a single density estimator and the less likely it is to be conservative. \Gls{nre} performs best at high levels of misspecification.
The \gls{sam} algorithms offer a substantial benefit to \gls{npe}, but is somewhat detrimental at high levels of misspecification for other approaches. We observe too that \gls{abc} is remarkably robust to misspecification, producing conservative posteriors at every level. Finally, it is notable that the ensemble is not always more conservative than the most conservative single estimator, nor is it always more conservative than the average coverage of a single estimator.

\subsection{Task: Stochastic Volatility}
\glsunset{sv}
\glsunset{svss}
A stochastic volatility model similar to an example used in \cite{hoffman2014no}\footnote{Our particular implementation closely follows \url{https://num.pyro.ai/en/stable/examples/stochastic_volatility.html}}. Consider the two-dimensional parameter $\bth=(\tau, \nu)$ with independent prior distributions $\tau \sim \mathrm{Gamma}(5, 25)$ and $\nu \sim \mathrm{Gamma}(5, 1)$. The generative process is then taken to be
\begin{equation}\label{eq:sv-generative}
    s_0 \sim \mathcal{N}(0,\tau^{-2}); \quad s_i \overset{iid}{\sim} \mathcal{N}(s_{i-1},\tau^{-2}), \, i \in \llbracket 1,100 \rrbracket; \quad \bx \sim \mathrm{StudentT}_\nu(0, \exp(\bs))
\end{equation}
where the $t$-distribution uses a location-scale parameterisation.
The misspecification transform was chosen to emulate a period of high market volatility, of the kind that occurred during ``Volmageddon'', a day-long spike in a commonly traded volatility index (the \emph{VIX}) occurring in February 2018 \citep[see e.g.][]{augustin2021volmageddon}. Using the notation $\mathcal{S}=\llbracket 50,65 \rrbracket$ and $\mathcal{S}'=\llbracket 1,100 \rrbracket \setminus \mathcal{S}$, the final simulator output is generated by the deterministic transform $
    T_{\sigma}(\bx_i) = \bx_i \mathds{1}\{i \in \mathcal{S}'\} + 5 \sigma \bx_i \mathds{1}\{i \in \mathcal{S}\}$ for misspecification levels $\sigma \in \{0, 1, 2, 3, 4\}$ with $\sigma=0$ representing no misspecification. This transform is equivalent to adjusting the scale of the Student-t distribution in Equation \eqref{eq:sv-generative}. The NUTS algorithm \citep{hoffman2014no}, as implemented in NumPyro \citep{bingham2019pyro}, was used to generate reference posterior samples. 

\begin{figure}[!htb]
  \centering
  \includegraphics[width=\linewidth]{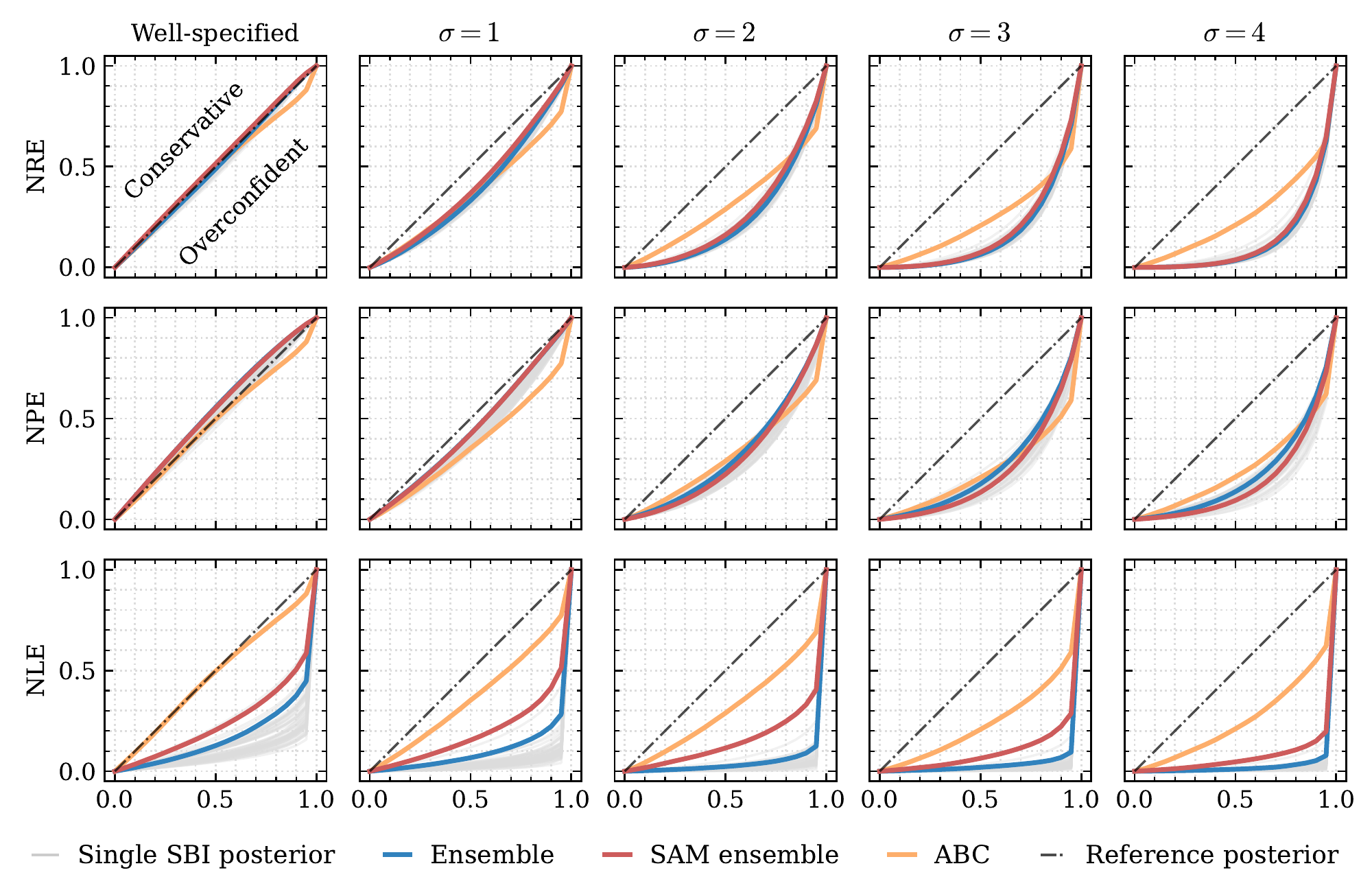}  
\caption{Coverage results for all algorithms on \gls{sv} task with $10^5$ training samples.}
\label{fig:sv-condensed}
\end{figure}

\paragraph{Results.} Figure \ref{fig:sv-condensed} shows the coverage results for the \gls{sv} model using $10^5$ training samples. Once again the performance of individual posterior approximations and the resulting ensembles all degrade substantially under increasing misspecification. In this example, the \gls{sam} training procedure makes a noticeable contribution only for \gls{nle}, which with or without the addition is unusable. \Gls{abc} does not perform conservatively, which is likely due to bias. Finally, unlike in the \gls{tgss} task, no algorithm offers satisfactory performance across all levels of misspecification.

\subsection{Task: SLCP}
\begin{figure}[!htb]
  \centering
  \includegraphics[width=\linewidth]{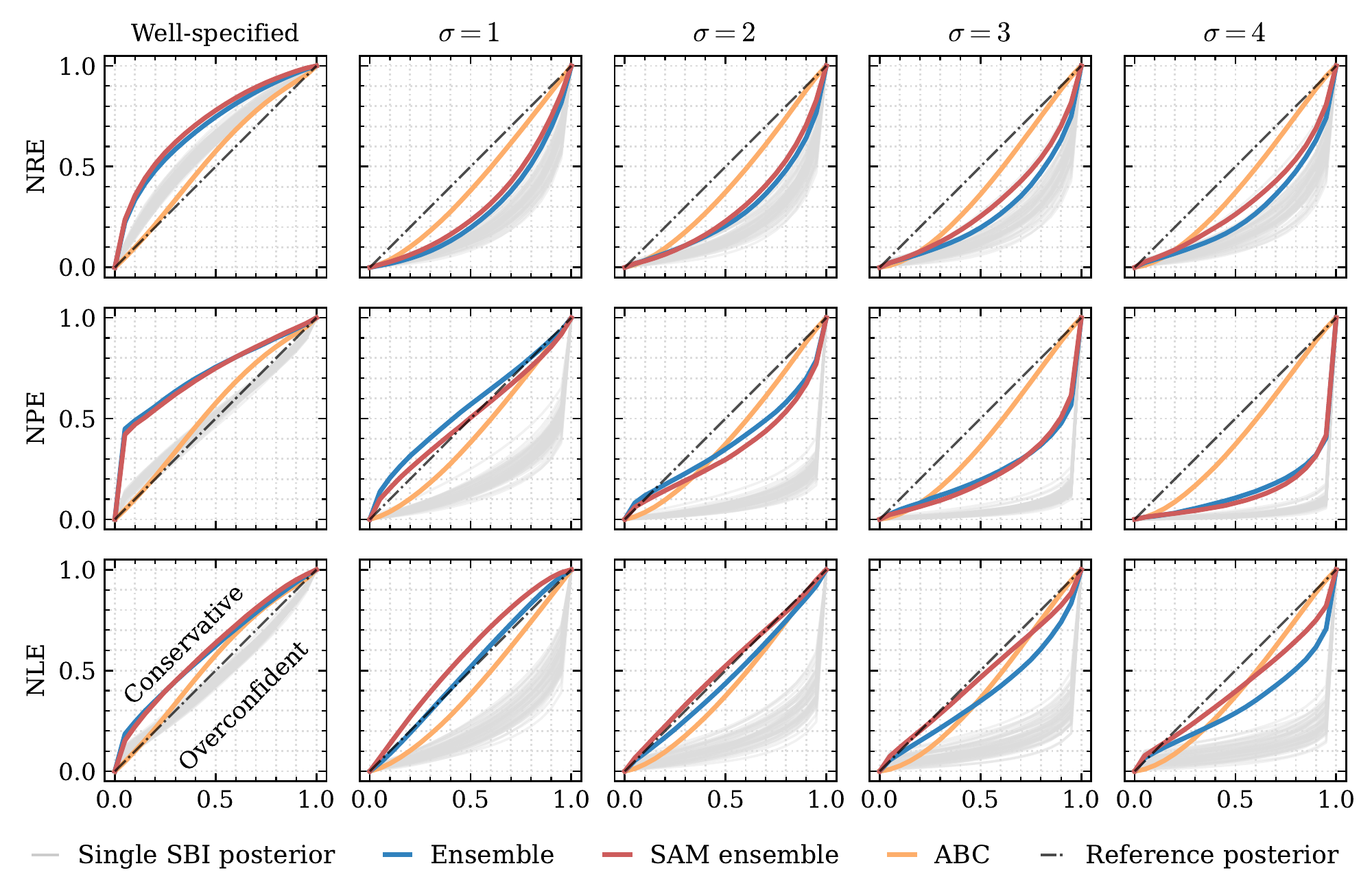}  
\caption{Coverage results for all algorithms on \gls{slcp} task with $10^4$ training samples.}
\label{fig:slcp-condensed}
\end{figure}

\glsunset{slcp}
A standard task in the \gls{sbi} literature \cite[e.g.][]{papamakarios2019sequential, lueckmann2021benchmarking}. The model is referred to here by the acronym \gls{slcp}, as it features a simple likelihood function and gives rise to a complex posterior. It has five-dimensional parameter $\bth$ with prior distribution \emph{iid} $\bth_i \sim \mathcal{U}(-3,3)$. The parameters define a mean and covariance matrix
\begin{equation}
    \mathbf{m}_{\bth} = (\theta_1, \theta_2) \quad \mathrm{and} \quad  \mathbf{S}_{\bth}=
    \begin{pmatrix} s_1^2 & \rho s_1 s_2 \\ \rho s_1 s_2 & s_2^2  \end{pmatrix} \quad \mathrm{with} \quad s_1=\theta_3^2, s_2 = \theta_4^2, \rho = \tanh(\theta_5)
\end{equation}
from which the model output $\bx = (\bx_1, \ldots, \bx_4)$ is drawn in an \emph{iid} fashion $\bx_i \sim \mathcal{N}(\mathbf{m}_{\bth}, \mathbf{S}_{\bth})$. Misspecification is added as a stochastic transform using additive Gaussian noise: $T_{\sigma}(\bx_i; \bz_i) = \bx_i + 100\sigma \bz_i$ where $\bz_i \sim \mathcal{N}(\bm{0}, (1 + \mathds{1}\{i=3\}) I_{2})$. Reference posterior samples were generated using a combination of Monte Carlo samplers as implemented in \texttt{sbibm} \citep[][]{lueckmann2021benchmarking}, a python library for benchmarking \gls{sbi} algorithms.
\paragraph{Results.} Figure \ref{fig:slcp-condensed} shows the coverage results for the \gls{slcp} model using $10^4$ training samples. In line with previous examples, individual posterior approximations increase in variance and become increasingly overconfident as misspecification increases. Here, \gls{sam} offers some benefit to \gls{nre} and \gls{nle}, but not \gls{npe}. Similar to the \gls{tgss} task, \gls{abc} is highly robust to misspecification, producing only slightly overconfident posteriors at the highest level of misspecification tested.

\section{Related Work}

\glsreset{abc}

\paragraph{Approximate Bayesian Computation.} 
\Gls{abc} is a widely used Monte Carlo method for approximating Bayesian posterior distributions in settings with intractable likelihoods. In its most basic form it can be thought of as a rejection Monte Carlo algorithm with a distance threshold controlling the desired fidelity of the approximation. In general, \gls{abc} requires more model samples than most \gls{sbi} methods to achieve similar results \citep{lueckmann2021benchmarking}, which is undesirable when the simulator is expensive to run. However, as it relies only on traditional Monte Carlo sampling strategies, it is (currently) far more amenable to theoretical study. In the present context, a valuable property of \gls{abc} is its relatively well understood performance under model misspecification \citep[][]{frazier2017model, frazierrobust, ridgway2017probably}.
Finally, we note that \gls{abc} does not exactly target the model posterior. It can in fact be interpreted as an exact method assuming a particular noise model \citep{wilkinson2013approximate}. It can also be cast as a generalised Bayesian posterior \citep{schmon2020generalized}, another relevant class of approximations which we now describe.

\paragraph{Generalised Bayesian Inference} 
\Gls{gbi} describes a framework encapsulating several methods that facilitate Bayes-like updating of Gibbs posteriors, that is, posteriors of the form $p_\ell(\bth \mid \by) \propto \exp(-\ell(\by;\bth))\pi(\bth)$ for some loss function $\ell$. One motivation for this is model misspecification, since in this setting it can be argued that updating one's beliefs according to the model likelihood is not a reasonable inferential choice. Such ideas can be approached from several directions, including PAC-Bayes \citep{guedj-pac}, coherent belief updating \citep{bissiri2016general}, and optimisation \citep{knoblauch2019generalized}. As with \gls{abc}, \gls{gbi} targets not the model posterior but another distribution altogether. 
Such ideas have been explored for likelihood-free models by \citet{schmon2020generalized, dyer2021approximate, pacchiardi2021generalized} and \citet{dellaporta22a}.
While this is of course epistemologically sound, our aim here is to ensure that the model posterior can be assessed accurately using neural \gls{sbi} techniques, even when the observation is unlikely under the assumed model.

\section{Conclusion}
\label{sec:conclusion}
In deploying \gls{sbi} methods to infer parameters of simulation models, practitioners reconcile simulation output with real data. However, real data can be of poor quality, corrupted by noise, or may simply not obey model assumptions.
This analysis is the first to demonstrate that if such deviation occurs, current state-of-the-art neural \gls{sbi} techniques can fail catastrophically. When we add relatively innocuous data transformations to simulated data, for example a period of higher volatility (volatility shock) in a stochastic volatility model, the estimated posteriors frequently fail to even cover the posterior mass of the true data posterior.
We demonstrated that this failure is not confined to a single method, but rather it appears systematically across all methods and tasks tested, and hence cannot be explained simply by the shortcomings of any single neural density estimation algorithm. It appears to be partially, though by no means wholly, mitigated in most settings through ensembling of posteriors, possibly trained using robust techniques, and avoidance of sequential schemes.

Our findings are concerning if neural \gls{sbi} techniques are to be relied on for scientific discovery or decision making. While we do not propose any generally applicable mitigation strategy in the current work, we hope the results of the benchmark presented offer a clear indication that more work is required to ensure robustness of neural \gls{sbi} algorithms in real world use.

\bibliographystyle{abbrvnat}
\bibliography{refs} %

\appendix

\section*{Appendix}

\section{Further Experimental Details}

To run the algorithms, we made use of the \texttt{sbi} package \citep{tejero-cantero2020sbi}. The \gls{slcp} example was adapted from the benchmarking library \texttt{sbibm} [\url{https://github.com/sbi-benchmark/sbibm}] accompanying the paper \cite{lueckmann2021benchmarking}. As noted in the main text, the SV example was adapted from \url{https://num.pyro.ai/en/stable/examples/stochastic_volatility.html}. Throughout the experiments, \texttt{PyTorch} \citep{pytorch_neurips} was used. Experiment config files were managed and launched using \texttt{hydra} \citep[][]{Yadan2019Hydra} with \texttt{joblib} \citep[][]{joblibref}. The \gls{sam} implementation was adapted from \url{https://github.com/davda54/sam}.

We now describe the algorithms used in greater detail. Unless otherwise stated, the \texttt{sbi} library default settings were used. Firstly, for all algorithms but \gls{abc}, i.e.~\gls{nre}, \gls{npe}, \gls{nle}, as well as their \gls{sam}-trained variants, the AdamW optimiser \citep[][]{loshchilov2017decoupled} was used with learning rate $\gamma = 3\times 10^{-4}$ (and standard \texttt{PyTorch} settings otherwise). We found this performed slightly more robustly than standard Adam. For \gls{nle} and \gls{npe}, neural spline flows were used \citep[][]{durkan2019neural} with five transformation layers, 64 hidden features and a batch size of 128. For \gls{nre} a multilayer perceptron was used with three hidden layers of 64 nodes, and a batch size of 128. Training was performed to loss convergence (no improvement after 20 epochs) on a validation set, with no limit on the number of epochs otherwise. In preparatory experiments, we found the number of transformations used in the normalising flows and the number of layers used in the classifier made little qualitative difference to the robustness properties of the algorithms. Finally we note that, in order to keep the investigation as focused as possible, no embedding networks were used. 

\section{Full Results}
Below, we display in full the results of our benchmark experiments for every task. Described in full in Section \ref{sec:experiments}, these are in brief:
\begin{itemize}
    \item a toy Gaussian model, both with and without summary statistics (\gls{tg} and \gls{tgss}),
    \item a stochastic volatility model, both with and without summary statistics (\gls{sv} and \gls{svss}), and
    \item the \gls{slcp} model.
\end{itemize}
The \gls{svss} task is identical to the \gls{sv} task, but with the addition of summary statistics, namely the time-series mean, standard deviation, median, and median absolute deviation.

In all, the experiments represent over 10,000 CPU hours of computation. The experiments were run on a 96-core N2 GCloud machine with 768GB memory, using only CPU. 

\begin{figure}
  \centering
  \includegraphics[width=.99\linewidth]{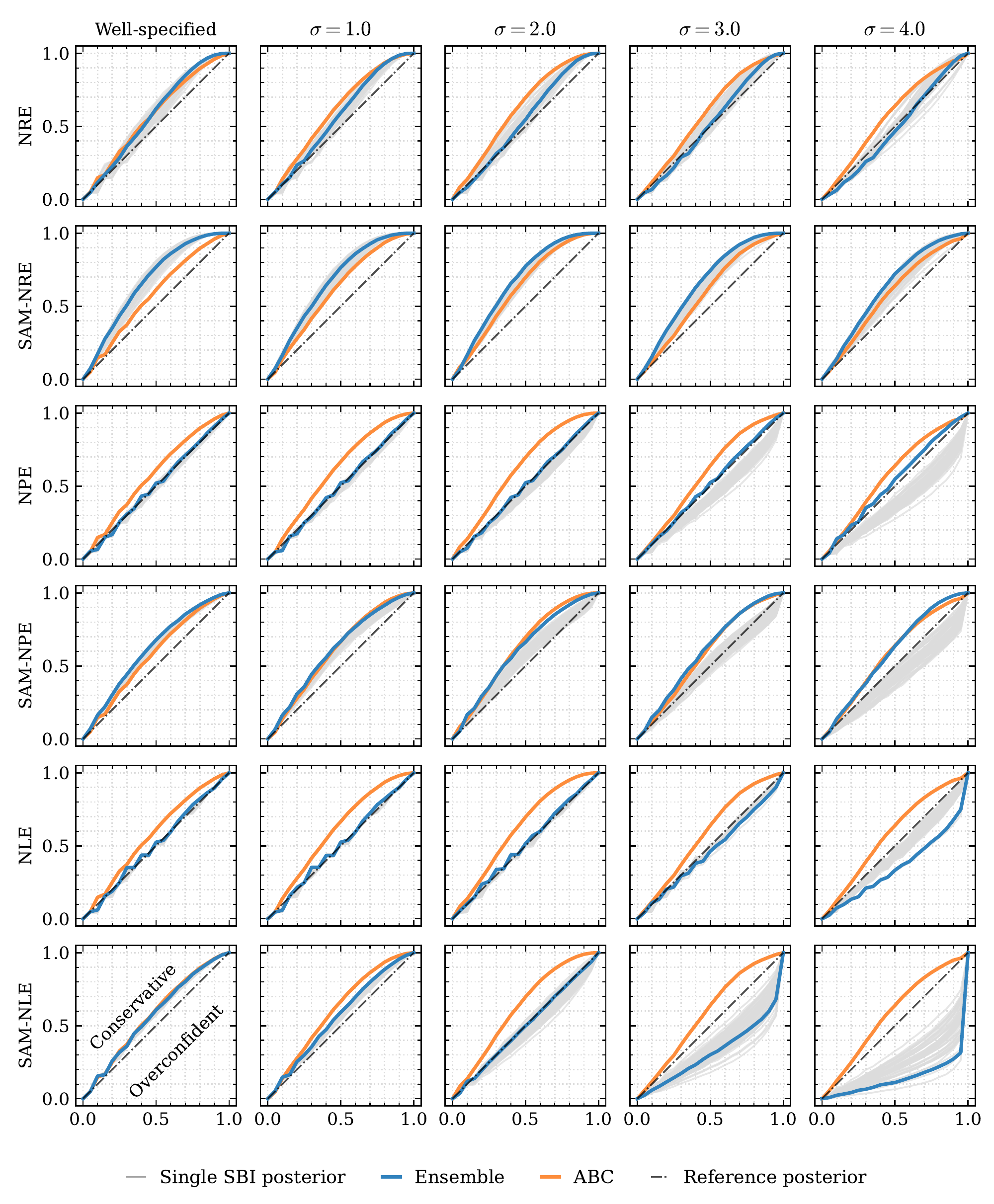}  
\caption{Coverage results for all algorithms on \gls{tg} task with $10^5$ training samples.}
\label{fig:tg-1e5}
\end{figure}

\begin{figure}
  \centering
  \includegraphics[width=.99\linewidth]{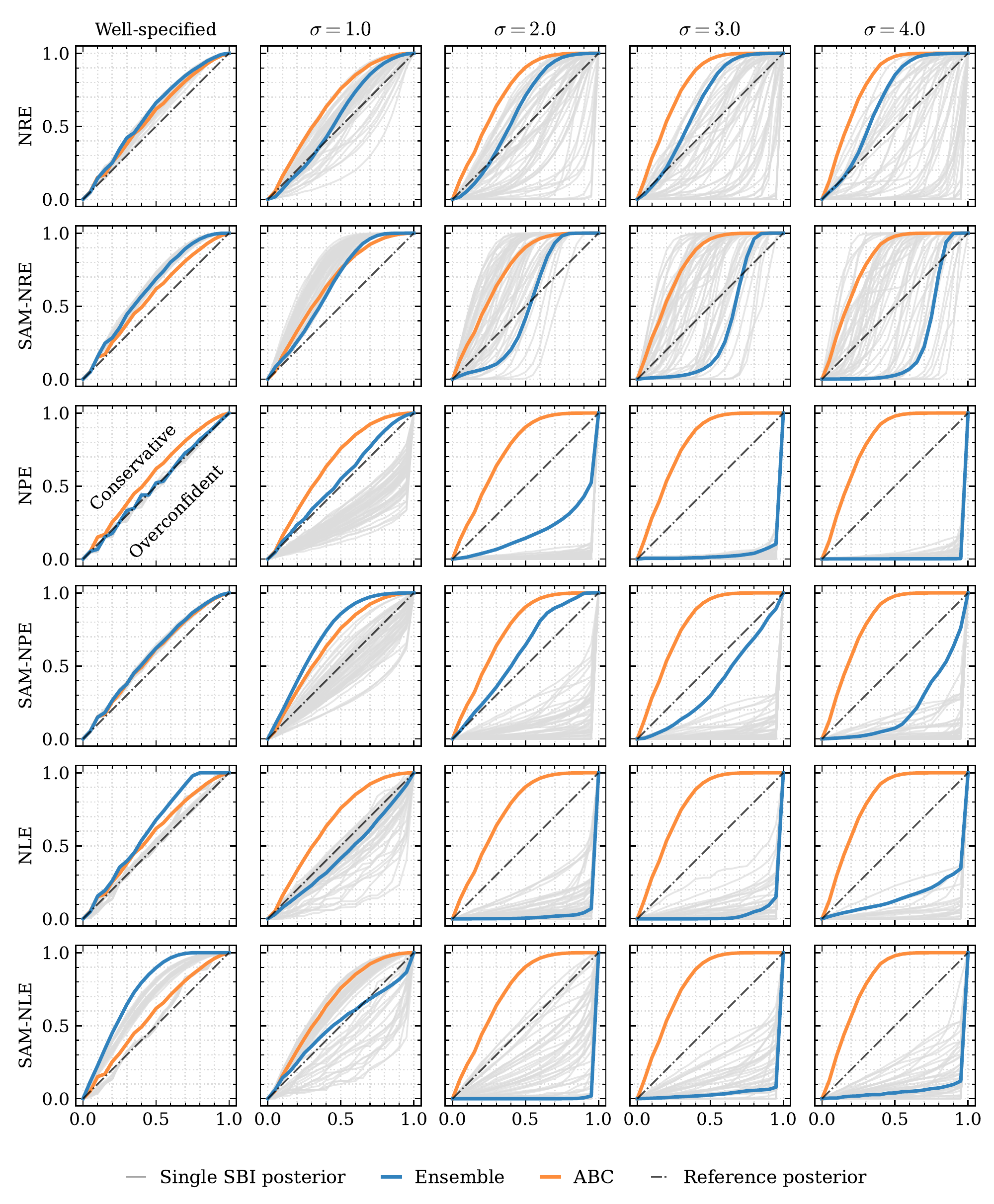}  
\caption{Coverage results for all algorithms on \gls{tgss} task with $10^5$ training samples.}
\label{fig:tgss-1e5}
\end{figure}

\paragraph{A note on the use of summary statistics} 
Our experiments demonstrate part of the complexity inherent in the use of summary statistics. As an example, consider the \gls{tg} model. The performance of every algorithm deteriorates when summary statistics are applied to the output of the model --- compare, e.g., Figures \ref{fig:tg-1e5} and \ref{fig:tgss-1e5}. It is not immediately obvious why this is the case. Indeed, at a first glance, the summary statistics used -- the sample mean and standard deviation -- are sufficient for the normal distribution. For this task, the mean alone is sufficient because the model variance is fixed, and the standard deviation is therefore an \emph{ancillary statistic}. It seems that the inclusion of a misspecified ancillary statistic (essentially acting as noise) alongside the sufficient statistic has degraded the performance of the posterior estimator.

\begin{figure}
  \centering
  \includegraphics[width=.99\linewidth]{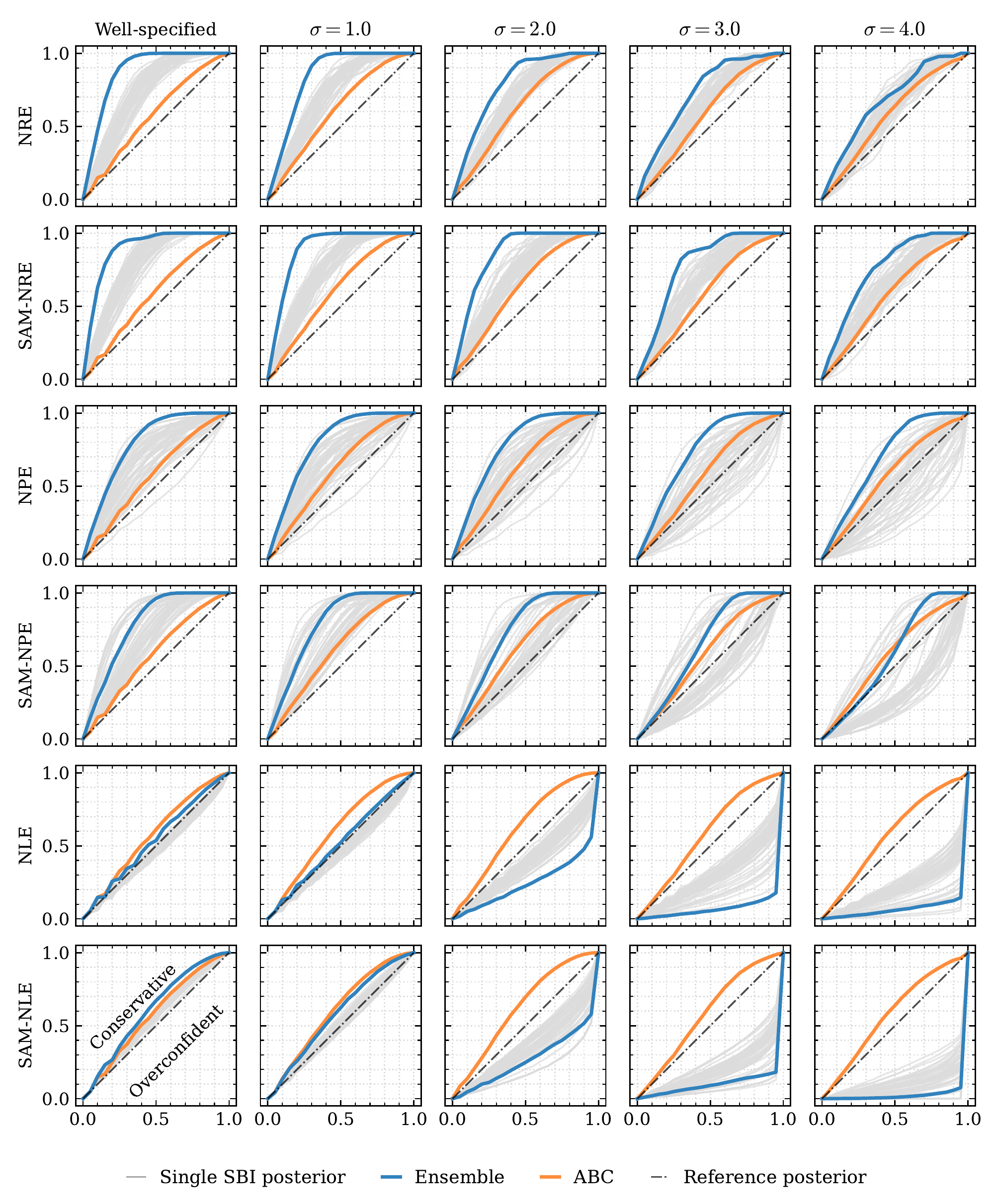}  
\caption{Coverage results for all algorithms on \gls{tg} task with $10^3$ training samples.}
\end{figure}
\begin{figure}
  \centering
  \includegraphics[width=.99\linewidth]{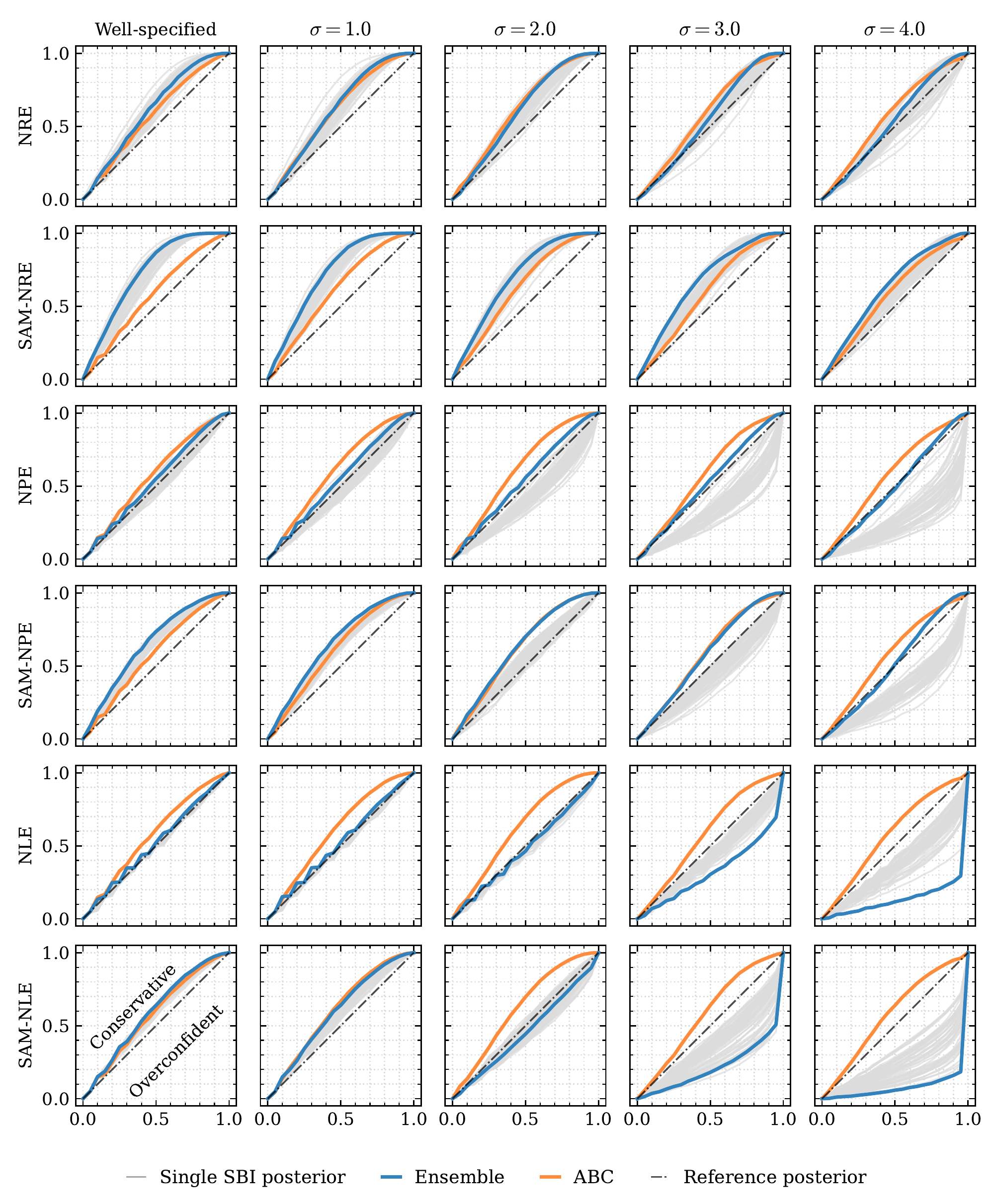}  
\caption{Coverage results for all algorithms on \gls{tg} task with $10^4$ training samples.}
\end{figure}

\begin{figure}
  \centering
  \includegraphics[width=.99\linewidth]{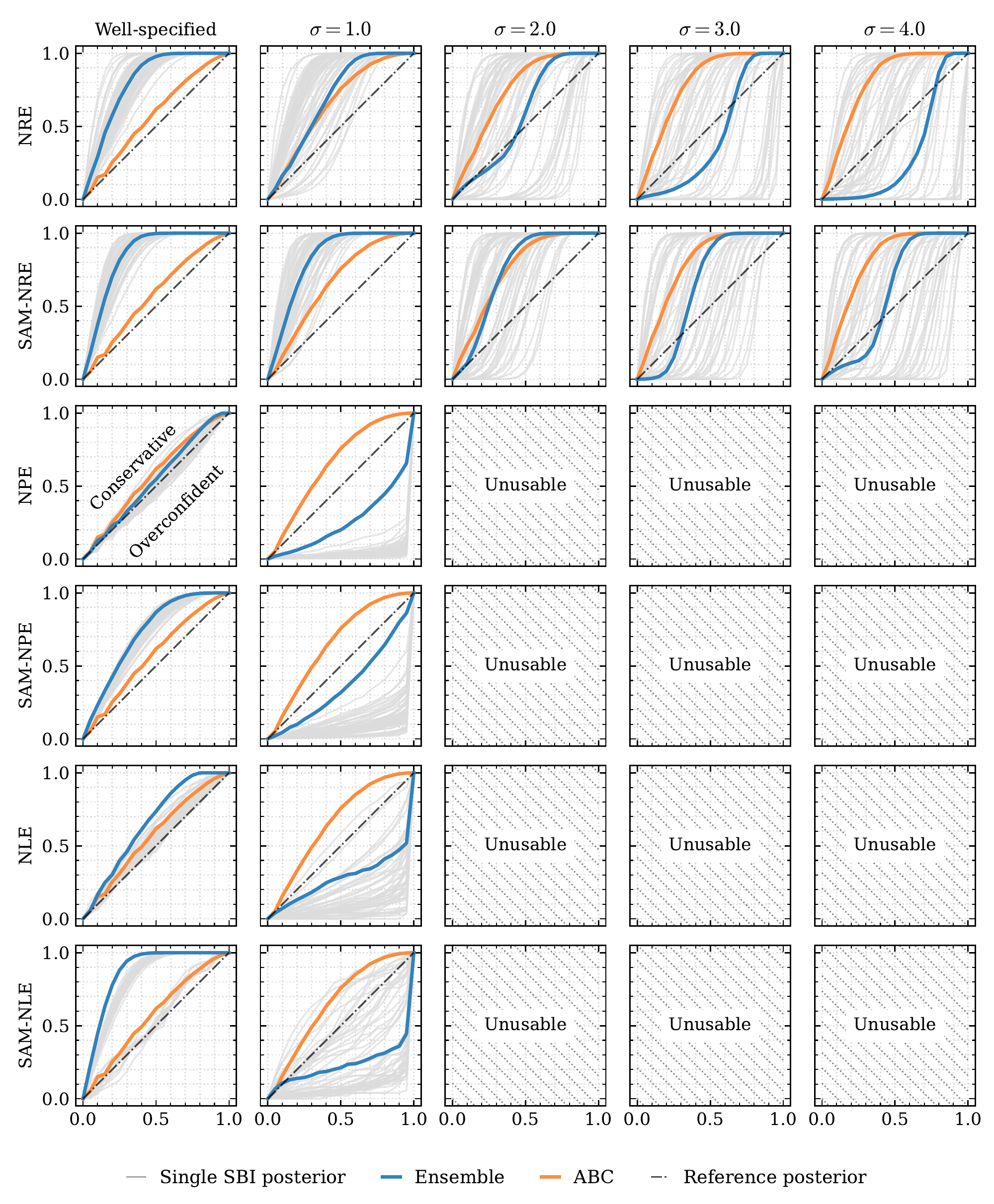}  
\caption{Coverage results for all algorithms on \gls{tgss} task with $10^3$ training samples.}
\end{figure}
\begin{figure}
  \centering
  \includegraphics[width=.99\linewidth]{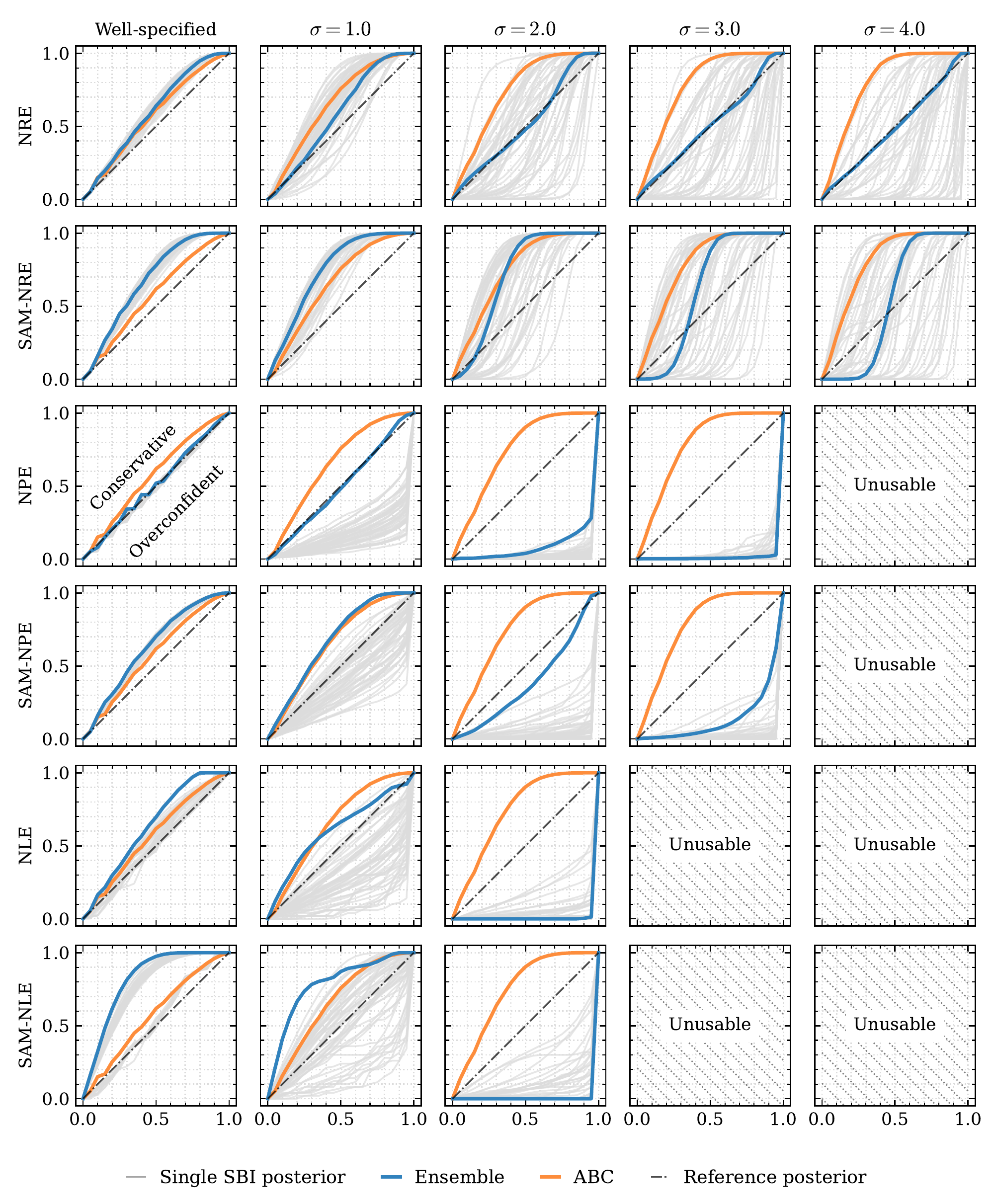}  
\caption{Coverage results for all algorithms on \gls{tgss} task with $10^4$ training samples.}
\end{figure}

\begin{figure}
  \centering
  \includegraphics[width=.99\linewidth]{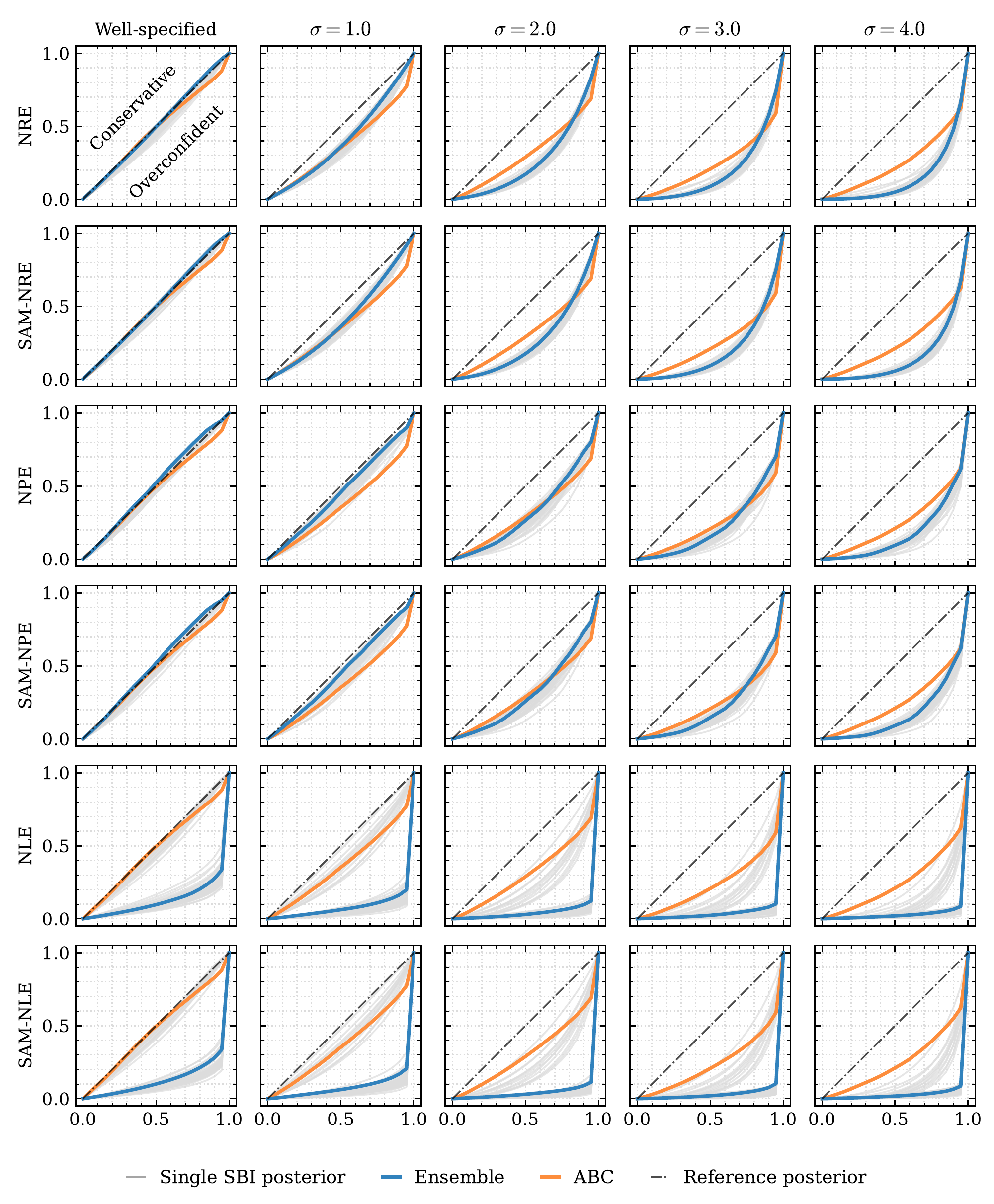}  
\caption{Coverage results for \gls{sv} task with $10^3$ training samples.}
\end{figure}
\begin{figure}
  \centering
  \includegraphics[width=.99\linewidth]{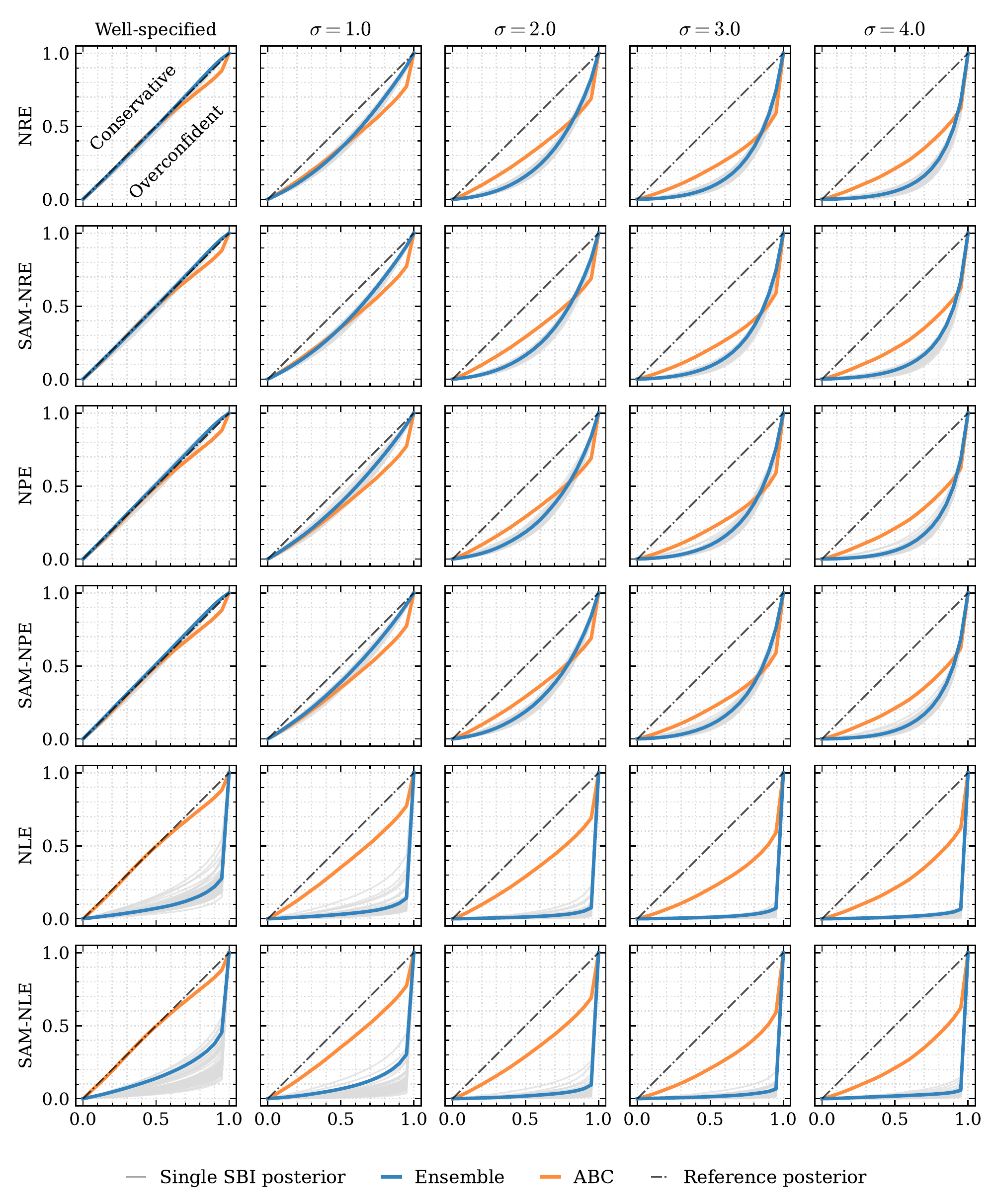}  
\caption{Coverage results for \gls{sv} task with $10^4$ training samples.}
\end{figure}
\begin{figure}
  \centering
  \includegraphics[width=.99\linewidth]{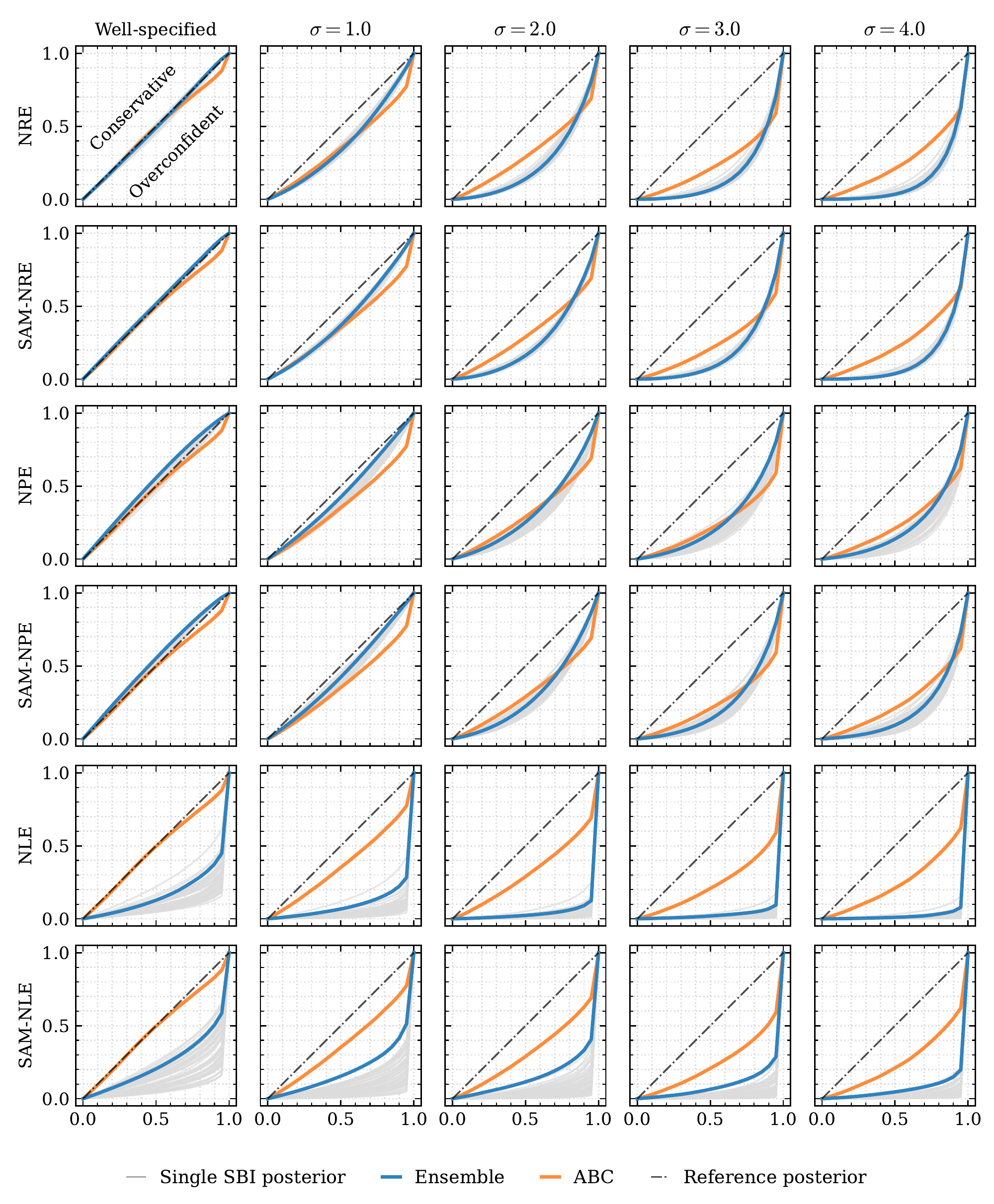}  
\caption{Coverage results for \gls{sv} task with $10^5$ training samples.}
\end{figure}

\begin{figure}
  \centering
  \includegraphics[width=.99\linewidth]{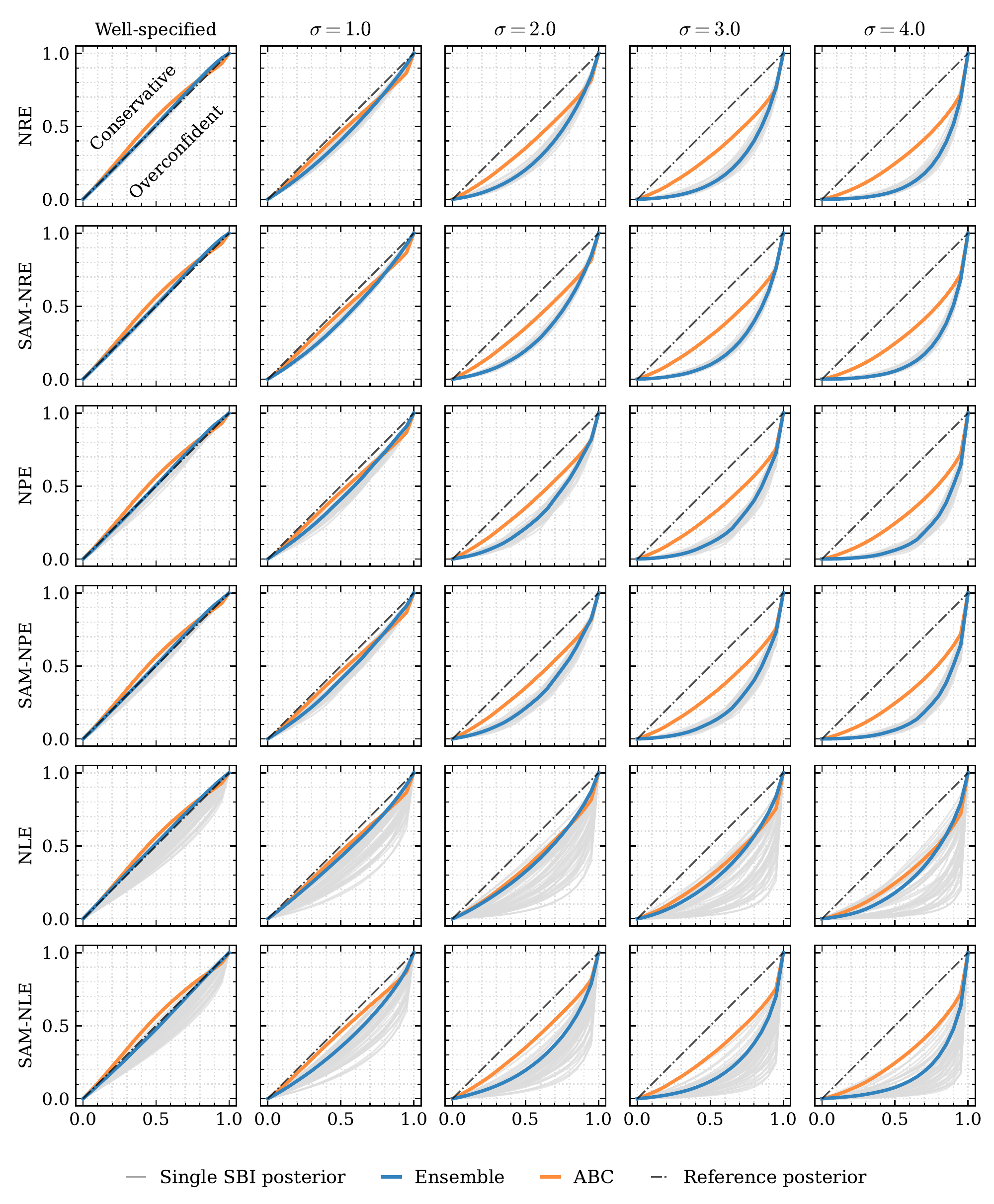}  
\caption{Coverage results for \gls{svss} task with $10^3$ training samples.}
\end{figure}
\begin{figure}
  \centering
  \includegraphics[width=.99\linewidth]{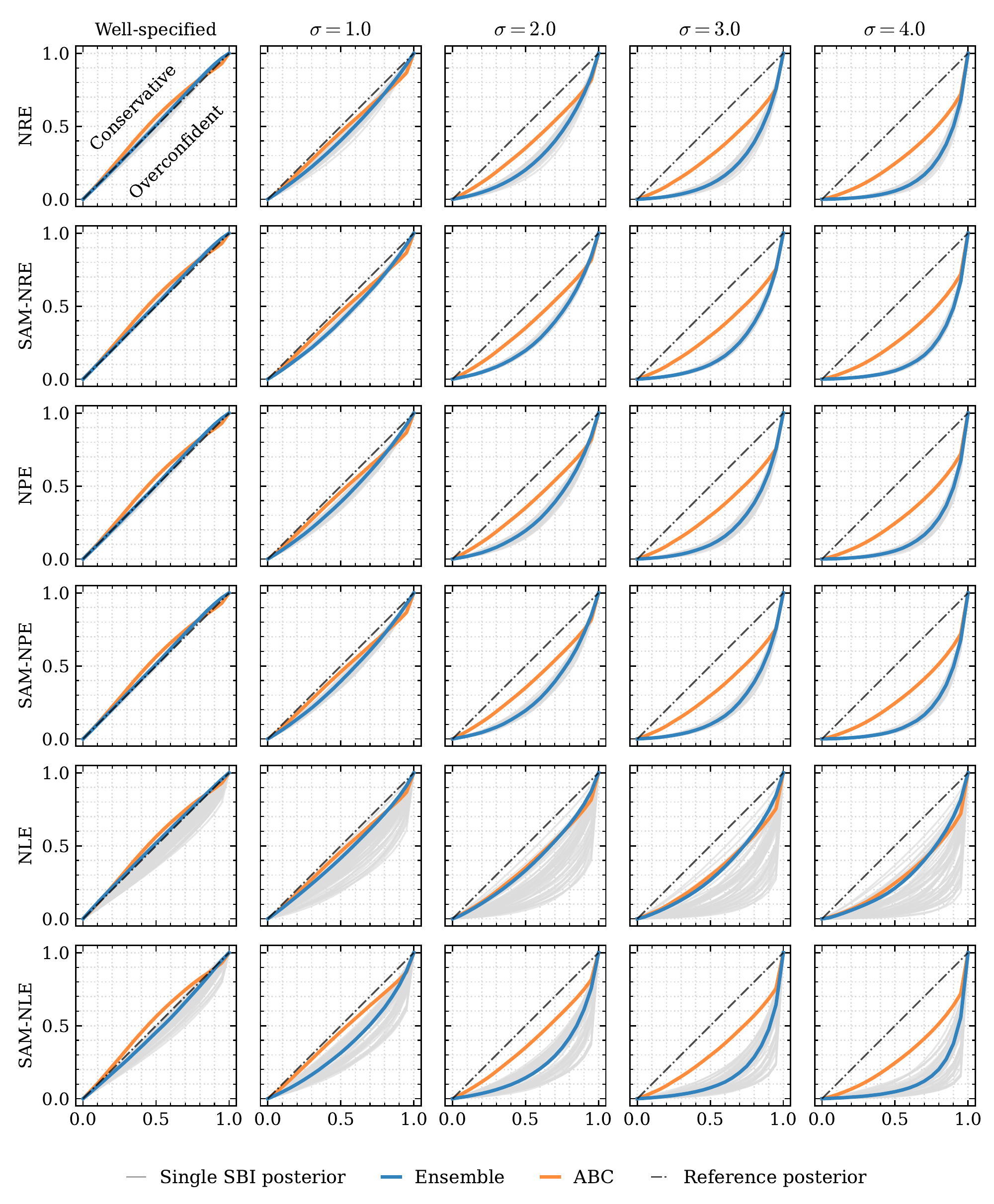}  
\caption{Coverage results for \gls{svss} task with $10^4$ training samples.}
\end{figure}
\begin{figure}
  \centering
  \includegraphics[width=.99\linewidth]{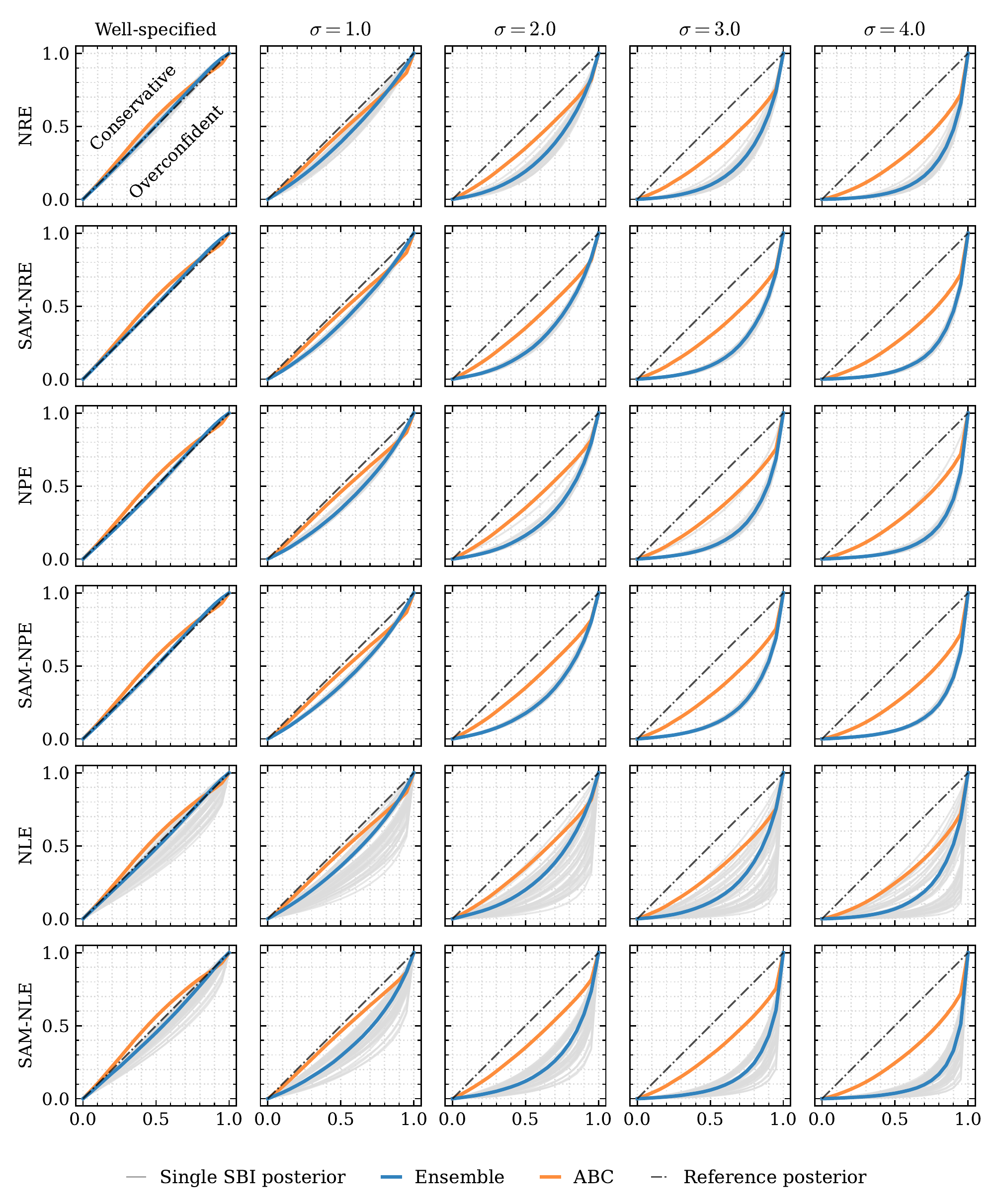}  
\caption{Coverage results for \gls{svss} task with $10^5$ training samples.}
\end{figure}

\begin{figure}
  \centering
  \includegraphics[width=.99\linewidth]{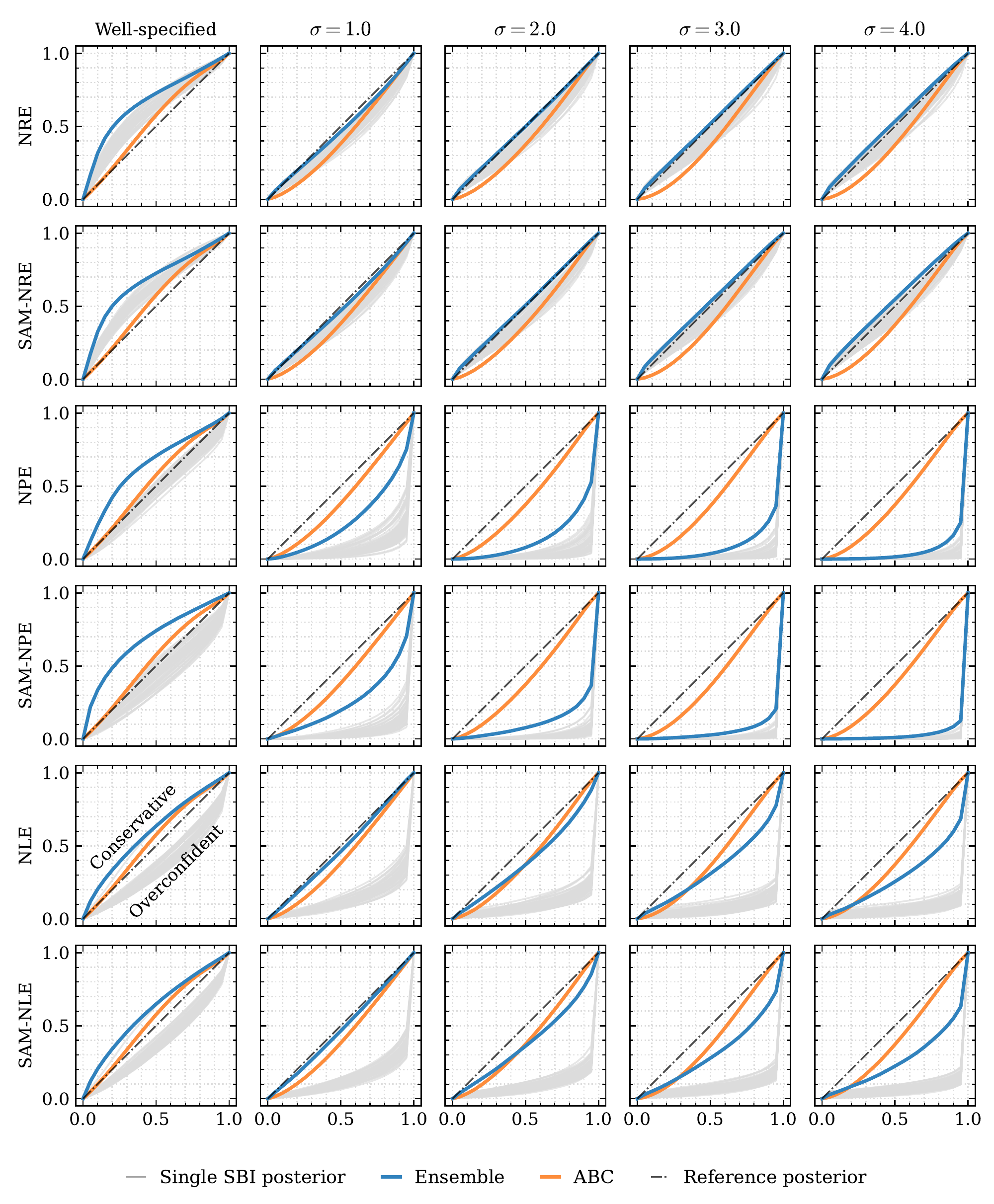}  
\caption{Coverage results for \gls{slcp} task with $10^3$ training samples.}
\end{figure}
\begin{figure}
  \centering
  \includegraphics[width=.99\linewidth]{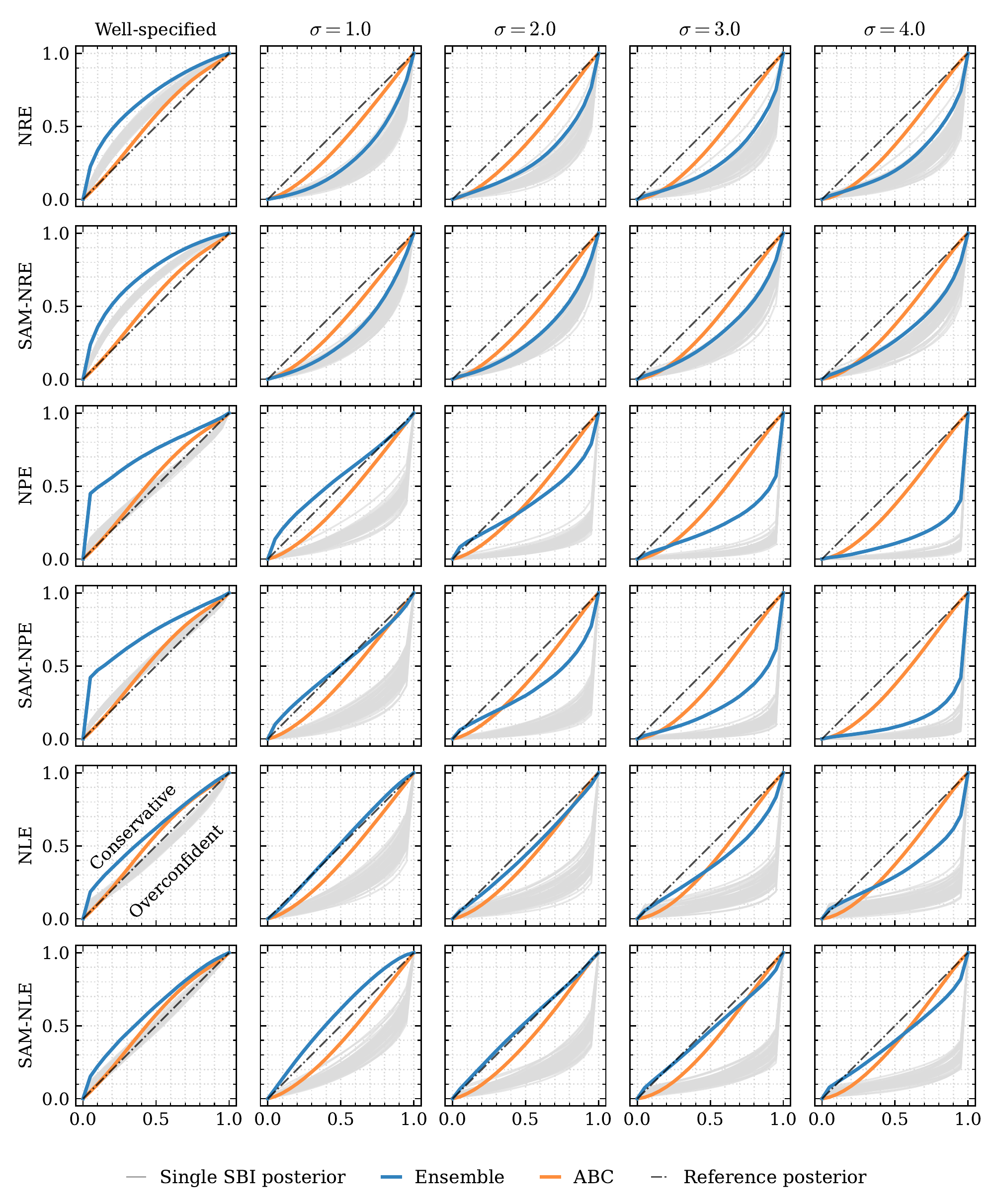}  
\caption{Coverage results for \gls{slcp} task with $10^4$ training samples.}
\end{figure}
\begin{figure}
  \centering
  \includegraphics[width=.99\linewidth]{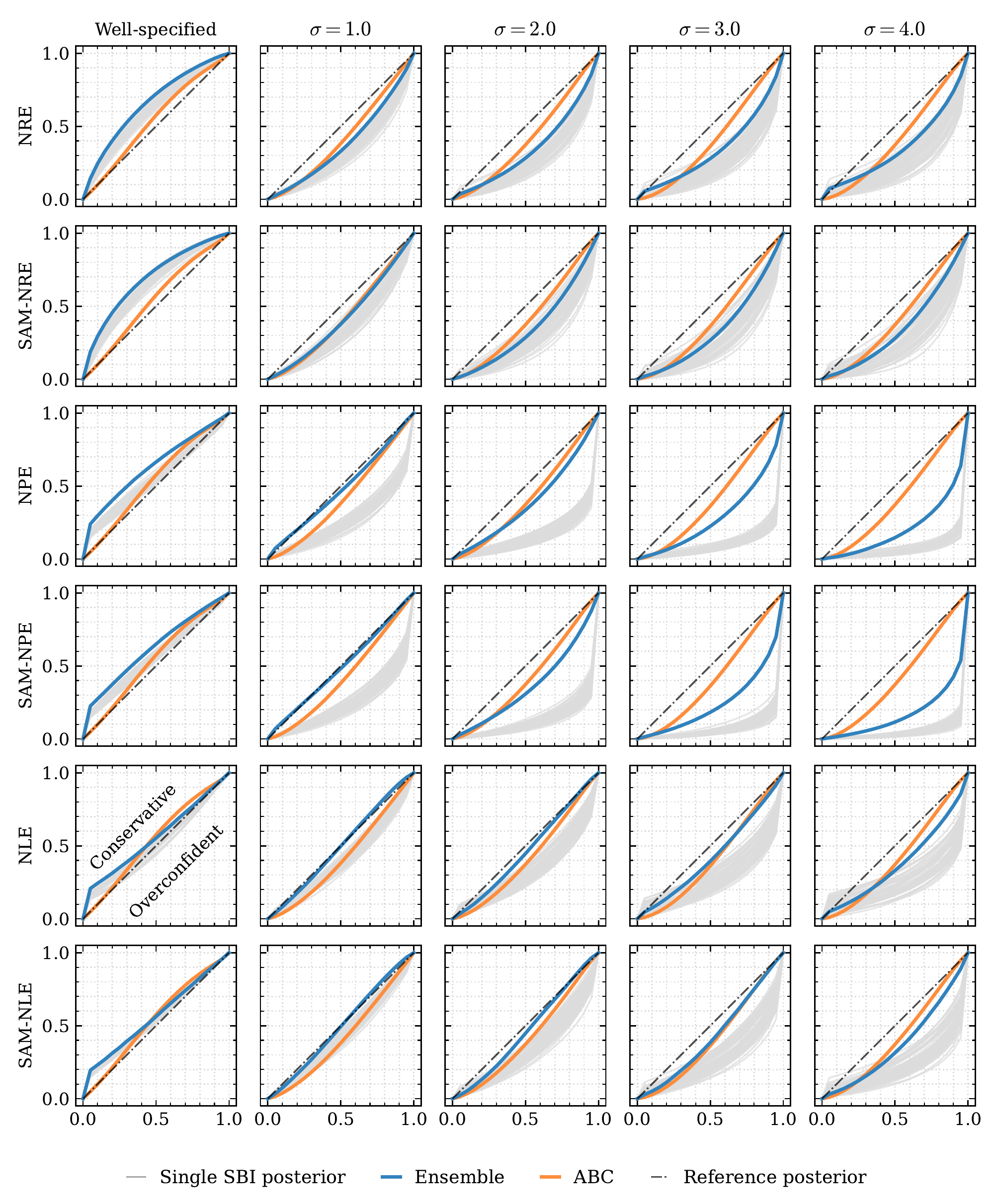}  
\caption{Coverage results for \gls{slcp} task with $10^5$ training samples.}
\end{figure}

\end{document}